\pdfoutput=1
\documentclass{article}

\usepackage{times}
\usepackage{graphicx} 
\usepackage{subfigure} 
\usepackage{multicol}
\usepackage{cuted}

\usepackage{natbib}

\usepackage{algorithm}
\usepackage{algorithmic}

\usepackage{amsmath,amsfonts,amssymb,url,array}

\usepackage{hyperref}

\graphicspath{home/db/Documents/UCL/matlab_code/code/}

\usepackage[accepted]{icml2015}

\icmltitlerunning{Learning share representations for value functions in Multi-Task RL}

\begin{document} 

\twocolumn[
\icmltitle{Learning Shared Representations for Value Functions in Multi-task Reinforcement Learning}

\icmlauthor{Diana Borsa}{diana.borsa@gmail.com}
\icmlauthor{Thore Graepel}{thore@google.com}
\icmlauthor{John Shawe-Taylor}{j.shawe-taylor@cs.ucl.ac.uk}
\icmladdress{University College London, Dept. of Computer Science, CSML, London WC1E 6EA, UK}

\icmlkeywords{reinforcement learning, representation learning, value functions, multi-task learning, transfer}

\vskip 0.3in
]

\begin{abstract} 
 We investigate a paradigm in multi-task reinforcement learning (MT-RL) in which an agent is placed in an environment and needs to learn to perform a series of tasks, within this space.  Since the environment does not change, there is potentially a lot of common ground amongst tasks and learning to solve them individually seems extremely wasteful. In this paper, we explicitly model and learn this shared structure as it arises in the state-action value space. We will show how one can jointly learn optimal value-functions by modifying the popular value-iteration and policy-iteration procedures to accommodate this shared representation assumption and leverage the power of multi-task supervised learning. Finally, we demonstrate that the proposed model and training procedures, are able to infer good value functions, even under low samples regimes. In addition to data
efficiency, we will show in our analysis, that learning abstractions of
the state space jointly across tasks leads to more robust, transferable representations with the
potential for better generalization.
\end{abstract} 

\section{Introduction}
\label{intro}
Reinforcement learning (RL) has gained a lot of popularity and has seen remarkable successes in the last years, exploiting and benefiting greatly from the recent developments in general functional approximators, such as neural networks \cite{Mnih2015}. At least part of this success seems to be linked to the ability of these universal functional approximators to distill meaningfully representations \cite{Bengio2009a}, from high-dimensional input states. These enabled RL to scale up to more complex environments and scenarios that were previously prohibited or required a great amount of feature engineering as shown in \cite{Mnih2015}, \cite{SilverHuangMaddisonEtAl2016}. Thus, learning a good abstraction of a given environment and the agent's role in it, seems to be a key component in developing complex and optimal control mechanisms.

While a lot of progress has been made in improving learning on individual single tasks, there seems to have been a lot less work in trying to re-use or efficiently transfer information from one task to the another \cite{TaylorStone2009}. Nevertheless, it is natural to assume that the different tasks an agent needs to learn  during its life, share a lot of structure and in-build redundancy. And potentially this could be leverage to speed-up learning.
 In this work we will propose a way to address this aspect, by learning robust, transferable abstractions of the environment that generalize over a set of tasks.  

Value functions are a central ideas in reinforcement learning \cite{Sutton1998}  and have been successfully used in conjunction with functional approximators to generalize over large state-action spaces. They are a concise way to readily assess the "goodness" of a state and can be learnt efficiently even in an off-policy fashion. This enables us to decouple the data gathering and the learning process, but most importantly this allows us to re-use past experiences collected under arbitrary or exploratory policies \cite{Sutton1998}. 
More recently, value functions have been shown to exhibit a very nice compositional structure with respect to the state space and goal states \cite{SchaulHorganGregorEtAl2015}. This is consistent with earlier studies in \cite{SuttonModayilDelpEtAl2011} that suggest value functions can capture and represent knowledge beyond their current goal that can be leveraged or re-used. Similar structures have been identified in the hierarchical reinforcement learning literature \cite{Dietterich2000} or \cite{SuttonPrecupSingh1999}. These all motivated our choice  of explicitly modelling the presence of this shared structure in the state-action value space.

 Using a multi-task RL formulation and following the recent work done in \cite{Calandriello2014}, we firstly outline two general ways of learning RL tasks jointly and sharing knowledge across them by extending two of the most popular procedures for learning value function, Fitted Q-Iteration \cite{Ernst2005} and  Fitted Policy Iteration \cite{AntosSzepesvariMunos2007}, to accommodate this shared structure assumption. Furthermore, taking advantage of the multi-task methods developed in supervised settings, we extend the work in \cite{Calandriello2014} to account for task-specific components.
 
 We will also show empirically that these lead to an overall improvement on the policies inferred, as well as a decrease in the number of samples per task needed to achieve good performance. We explore the nature of the representation learnt and its potential transferability to new, but related tasks. We show this learning is able infer a compressed structure that nevertheless captures a lot of transferable knowledge, similar to option-like transition models \cite{SuttonPrecupSingh1999} -- without us ever specifying a partition of desirable states or subgoals.
  Finally we will argue that this way of learning, leads to more robust and refined representations which are deemed crucial for learning and planning in complex environments.

\section{Proposed Model}
\subsection{Background and Notation}
We define a Markov Decision Process (MDP) as a tuple $\mathcal{M} = (\mathcal{S}, \mathcal{A}, \mathcal{P}, \mathcal{R}, \gamma)$, where $\mathcal{S}$ is the set of states, $\mathcal{A}$ is the set of actions\footnote{in this work this will be a finite set}, $\mathcal{P}: \mathcal{S}\times(\mathcal{S}\times\mathcal{A}) \rightarrow [0,1]$ is the transition dynamics $\mathcal{P}(s'|s,a)$ which provides a probability over next state $s'$, $\mathcal{R}:\mathcal{S}\times\mathcal{A} \rightarrow \mathbb{R}$ is a reward signal, which is assumed to be bounded ($\exists R_{max}$, s.t. $R(s,a) \leq R_{max}, \forall s \in \mathcal{S}, a \in \mathcal{A}$)  and $\gamma \in [0,1]$ is a discount factor.
\par Given an MDP and any policy $\pi:\mathcal{S}\times \mathcal{A} \rightarrow [0,1]$, we define the (state-action) value function, $Q^\pi(s,a)$ as the discounted cumulative reward an agent is expected to collect when starting from state $s\in\mathcal{S}$, taking action $a\in\mathcal{A}$ and then act accordingly to policy $\pi$:
\begin{equation}
Q^{\pi} (s,a) = \mathbb{E}_{\pi, \mathcal{P}} \left[ \sum_{t=0}^{\infty}{\gamma^t r_t} | s=s_0, a=a_0\right]
\end{equation}
The expectation is over all trajectories starting in $(s,a)$ and obtained by interacting with the environment ($\mathcal{P}$) while following behaviour policy $\pi$.
\par Our goal is to learn an optimal behaviour with respect to this expected cumulative reward. Thus we are looking for $\pi^*$ s.t.
\begin{equation}
\pi^*(s,a) = \arg \max{_\pi}{Q^\pi(s,a)}
\end{equation}
We will denote this optimal value function as $Q^* = Q^{\pi^*}$.  
And note that finding $Q^*$, automatically gives us an optimal policy $\pi^*$ by acting greedily with respect to these values. In the following, we denote this greedy operation by  $\pi^*= \mathcal{G} Q^*$.
\subsection{Problem formulation}
We will consider the scenario in which an agent resides (or is placed) into an environment in which it needs to perform a series of tasks. The overall goal  is to learn how to succeed at all these tasks. The environment is described by a state-action space $S \times A$ and a transition kernel $\mathcal{P}(s'|s,a)$ and the tasks can be specified by different rewards signals $\mathcal{R}_t(s,a)$, one for each task $t = \overline{1,T}$. This formally gives rise to $T$ MDPs
$\mathcal{M}_t = (\mathcal{S}, \mathcal{A}, \mathcal{P}, \mathcal{R}_t, \gamma)$
which share a lot of structure. Thus, if we can find a way to leverage this structure, we expect this to aid the learning process and lead to better generalization.
  \cite{Taylor2009}
\subsection{A shared (value function) representation}
We propose to model the shared structured found in above defined MDP-s as a shared embedding of the state-action space $\phi: \mathcal{S} \times \mathcal{A} \rightarrow \mathbb{R}^d$ on which we can build the individual optimal value functions $\{ Q^*_t\}_{t=1,T}$ for all considered tasks and potentially new ones.
\begin{figure*}
	\centering
	\includegraphics[width = 0.85\textwidth]{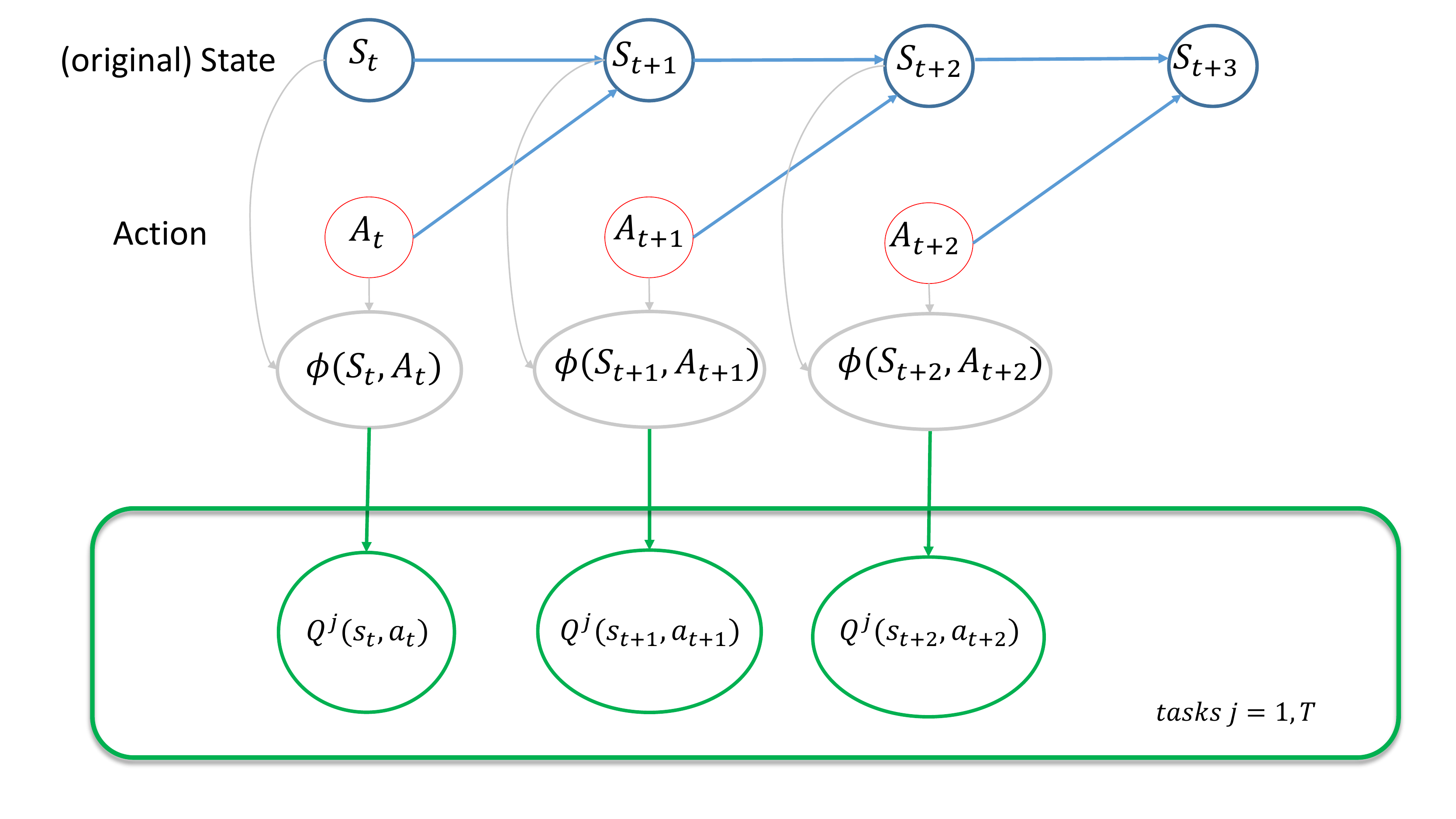}
	\caption{Proposed Model: Enforcing a shared representation of the state-action space, to be used in modelling all value functions, $\{Q_j\}$ across a set of tasks $j=1,T$. }
	\label{MT_Qlearning}
\end{figure*}
Thus in this paper we are interested in learning this shared embedding as well as ultimately the optimal behaviour for each of the tasks considered. In the following, we will present how one can extend two of the most popular paradigms of learning value functions, Fitted Q-Iteration and Fitted Policy Iteration, to incorporate this shared structure assumption. This will come down to employing a multi-task learning procedure in the target-fitting step of Q-Iteration and in the policy evaluation step of Policy-Iteration. 
\section{Multi-task (Fitted) Value Iteration}
In this section we outline a general framework of using \textit{approximate value iteration} to infer the optimal $Q$-values (and optimal policies) for a set of tasks, in a given environment following the MT-RL setup previously introduced. The proposed algorithm is an extension of Fitted Q-Iteration (FQI) that allows for joint learning and transfer across tasks. Following the recipe of FQI, at each step in the iteration loop and for each sample in our experience set $\mathcal{D} = \{(s,a,r,s')|s'\sim \mathcal{P}(.|s,a)\}$, we compute the one-step TD target based on our current estimate of the value function. Then, treating these estimates as ground truth, we obtain a regression problem from the state-action space onto the TD targets (which are really place-holders for the true value function).  In the case of MT-RL, we will obtain such a regression problem for each task $t$. Now we could, in principle, solve all these regression problems independently for each task, which would amount to applying FQI individually to each task. But our assumption is that there is shared structure between tasks and we would like to make use of this common ground to aid the learning process and arrive at more robust abstractions of the input space. Thus we propose solving the regression problems jointly, accounting for and building upon a common representation. A detailed description of the proposed procedure is outlined in Algorithm \ref{alg:MT-FQI}.

\begin{algorithm}                      
\caption{Multi-task Fitted Q-Iteration}     
\label{alg:MT-FQI}                           
\begin{algorithmic}                    
\REQUIRE $\mathcal{D} = \cup_{t=1}^T \mathcal{D}_t \sim \mu, \mathcal{P}$ - set of experiences/episodes for each task $t$
\vspace{1em}
\STATE \textbf{Initialize} $\Theta = \Theta_0, k = 0$ (parameters)
\vspace{1em}
\WHILE{not converged $(d\Theta< \epsilon || k< \text{MaxIter})$}
\STATE \textbf{Compute Targets}:
 \\ $ \mathcal{Y}_t^{(k+1)} = \{ y_t^{(k+1)}(s,a) = R_t(s,a) + \gamma \max_{a'}Q^{(k)}(s',a')|(s,a,s') \in \mathcal{D}_t \}, \forall t = 1,T$ \\

\STATE \textbf{Multi-task Learning:}  
\\ $\Theta^{(k+1)} = \text{MTL}(\mathcal{D} = \cup_{t=1}^T \mathcal{D}_t, \mathcal{Y}^{(k)} = \cup_{t=1}^T \mathcal{Y}_t^{(k)})$ 
 \\
 
 \STATE  $d\Theta = \Vert \Theta^{(k+1)} - \Theta^{(k)} \Vert, k=k+1$ 
\ENDWHILE
\vspace{1em}
\STATE \textbf{Return: \\} $\Theta = \{\theta_t\}_{t=1}^T \left( \Rightarrow Q_{t}(s,a) = f_{\theta_t}(s,a), \forall s,a \in \mathcal{S}\times \mathcal{A} \right) $
\end{algorithmic}
\end{algorithm}
Note that, in the spirit of generalization, we do not specify a particular algorithm for the multi-task learning step(MTL in Algorithm  \ref{alg:MT-FQI}). There is extensive literature of how to deal with multi-task inference and exploit shared structure between tasks in purely supervised settings, and we will take a look at some instantiations of this step throughout this work.

%
%

\section{Multi-task (Fitted) Policy Iteration}
By a similar argument as the one presented in the last section for MT-FQI, we can extend the framework of \textit{general policy iteration} to the MT-RL scenario. Policy Iteration algorithms rely on an alternating procedure between a policy evaluation step and a policy improvement step. We can extend this framework to the multi-task case, by defining a (current) set of policies $\Pi = \{\pi_t\}_{t=1}^T$, one for each task, and then we evolve this set of policies jointly at each iteration, $k$. Please find an outline of the proposed procedure in Algorithm \ref{alg:MT-PI}. For now, we implement the policy improvement step by acting greedily with respect to our current estimates of the value function. This step is done individually for each task. 

\begin{algorithm}                      
\caption{Multi-Task Policy Iteration (MT-PI)}          
\label{alg:MT-PI}                           
\begin{algorithmic}                    
\REQUIRE $\mathcal{D} = \cup_{t=1}^T \mathcal{D}_t$, set of experiences for each task $t$
\vspace{1em}
 
\STATE \textbf{Initialize} $\Theta = \Theta_0, k = 0$
\vspace{1em}

\WHILE{convergence not reached $(d\Theta< \epsilon )$}
\STATE \textbf{Policy Evaluation}:
 \\ Compute $\Theta^{(k)} =$ MT-PE($ \mathcal{D}, \Pi^{(k)}$) via Algorithm \ref{alg:MT-PE} \\ where  $\Pi^{(k)} = \left[ {\pi_1}, \cdots,{\pi_t}, \cdots, {\pi_T} \right]$ and $\Theta^{(k)} = \left[ \theta^{\pi_1}_1, \cdots, \theta^{\pi_t}_t, \cdots, \theta^{\pi_T}_T \right]$

\STATE \textbf{Policy Improvement} 
 \\ $\pi_j^{(k+1)}(a|s) = \arg \max_a{Q^{\pi^{(k)}_t}_{t}(s,a)} = \arg \max_a{f_{\theta^{\pi_t}_t}(s,a)}$

\ENDWHILE

\vspace{1em}
\STATE \textbf{Return:} $\Theta = \{\theta_t\}_{t=1}^T$ and policies $\Pi = \{\pi_t\}_{t=1}^T \approx \Pi^*$
\end{algorithmic}
\end{algorithm}
On the other hand, we allow joint learning and sharing of knowledge in the policy evaluation step. This gives rise to a general procedure we will call Multi-task Policy Evaluation (MT-PE)- see Algorithm \ref{alg:MT-PE}.  In MT-PE, we are given a set of policies $\Pi = \{\pi_t\}_{t=1}^T$, one for each task, and a collection of experiences $\mathcal{D}= \cup_{t=1}^T \mathcal{D}_t$. Then the aim of the algorithm is to approximate the corresponding value functions $Q^{\pi_t}_t$ associated with acting out policy $\pi_t$ for task $t$. Note that, in general, this step (policy evaluation) requires on-policy data, for each policy $\pi_t$ and for each task $t$. This could be quite demanding and inefficient data-wise, as the numbers of tasks grow, not to mention that this is just the inner loop of another iterative algorithm (MT-PI). In this work, we opt for an implementation of the policy evaluation step that circumvents this problem. Making use of the Bellman Expectation Equation, we can compute regression targets for approximating $\{Q^{\pi_t}_t\}$ by only using experience of the form $(s,a,r_t,s')$ previously collected, as we did in the case of Fitted-Q Iteration.
\begin{equation}
\forall (s,a,r_t,s') \in\mathcal{D}_t \rightarrow y_t^{(i)} = r_t(s,a) + \gamma Q_t^{(i)}(s',\pi_t(s'))
\end{equation}
 Therefore, we have now reduced the original problem to a set of regression problems that can be solved jointly, under a shared input space representation. This is very similar to the multi-task learning step employed in MT-FQI, but now the shared structure is learnt to model the input set of policies $\Pi = \{\pi_t\}_{t=1}^T$, rather than the optimal ones. Nevertheless, by constantly improving the set of policies that are presented to the MT-PE step, we should eventually be able to convergence to the optimal policies and thus at this point, the policy evaluation step should be able to recover the shared structure amongst optimal value functions.

\begin{algorithm}                      
\caption{Multi-Task Policy Evaluation (MT-PE)}          
\label{alg:MT-PE}                           
\begin{algorithmic}                    
\REQUIRE $\mathcal{D} = \cup_{t=1}^T \mathcal{D}_t$, set of experiences for each task $t$ 
\\ A set of policies $\Pi =  \{\pi_t\}_{t=1}^T$, for each task $t$ that need to be evaluated
\vspace{1em}

\STATE \textbf{Initialize} $\Theta = \Theta_0, i = 0$
\vspace{1em}

\WHILE{convergence not reached $(d\Theta< \epsilon)$}
\STATE \textbf{Compute Targets}:
 \\ $ \mathcal{Y}_t^{(i+1)} = \{ y_t^{(i+1)}(s,a) = R_t(s,a) + \gamma Q^{(i)}_t(s',\pi_t(s'))|(s,a,s') \in \mathcal{D}_t \}, \forall t = 1,T$ \\

\STATE \textbf{Multi-task Learning:}  
 \\ $\Theta^{(i+1)} = \text{MTL}(\mathcal{D} = \cup_{t=1}^T \mathcal{D}_t, \mathcal{Y}^{(i)} = \cup_{t=1}^T \mathcal{Y}_t^{(i)})$ 
 \\
 
 \STATE  $d\Theta = \Vert \Theta^{(i+1)} - \Theta^{(i)} \Vert, i=i+1$ 
\ENDWHILE

\vspace{1em}
\STATE \textbf{Return:} $\Theta = \{\theta_t\}_{t=1}^T$ \\ $\left( \Rightarrow \hat{Q}_{t}^{\pi_t}(s,a) = f_{\theta_t}(s,a) \approx Q_{t}^{\pi_t}(s,a), \forall s,a \in \mathcal{S}\times \mathcal{A} \right) $
\end{algorithmic}
\end{algorithm}

\section{Multi-task and Representation Learning}
In this section will we look at a couple of methods we can plug into the above algorithms in the $MTL$ step. For this we will assume a linear parametrization of the state-action value space -i.e. we assume $\exists \Phi = \{\phi_k: \mathcal{S} \times \mathcal{A} \rightarrow \mathbb{R} \}$ s.t. all value function of interest ${Q_t}$ can be well approximated by this linear combination of this set of features. In the case of fitted value iteration we want this set of features to fit well the intermediate targets $\mathcal{Y}^{(i)}$, but ultimately we are interested in a set of features that fit well the optimal value functions $Q^*_t$ and we will see that this turns out to be very small subspace of the original feature space.

\begin{equation}
Q^{*}_t(s,a) = \langle \Phi^*(s,a), w^{}_t \rangle, \forall t = \overline{1,T}
\label{eq:linear_parametrization_value_function}
\end{equation}

In the case of policy iteration, at each evaluation step we are interested in having a feature space that well approximates the value function corresponding to our current policies $\Pi^{(k)}$. Thus we are looking for $\Phi_{PE}$ s.t. $Q^{\pi_t^{(k)}}_t \approx \langle \Phi_{PE}(s,a), w^{\pi_t^{(k)}}_t \rangle, \forall t = \overline{1,T}$. As policies improve, we will end up fitting optimal or near-optimal value functions. Certainly if the regression step can be done perfectly (no approximation error), policy iteration will continue to improve the policies and in the limit will converge to the optimal value functions. Thus, the representation that will come out of this learning procedure should be similar to the ones learned by value-iteration procedures. Consequently, ultimately what we want in terms of representation is a (low-dimensional) features space that spans the optimal value functions of interest.

\subsection{Multi-task feature learning}
In terms of planning, the joint problem we are trying to solve can be formalized as inferring
$\lbrace w_t \rbrace_{t=1}^T = \arg \min_{W} {\left[\sum_{t}{\mathcal{L}\left(w_t^T\Phi , \hat{Q}_{t} \right)} + \mathcal{H}(W)\right] }
$
where $\mathcal{H}(W)$ is a regularizer on the weight vectors, that encourages feature sharing. At the same time, 
we wish to learn a more compact abstraction of the state-action space, that will be shared among tasks. To make this a bit more formal, let $Q_{t, w}:\mathcal{S}\times\mathcal{A}\rightarrow \mathbb{R}, Q_{t, w}(s,a) := \langle \Phi(s,a), w_t \rangle $, then our assumption can be expressed as: $\exists$ a small set of features $ \{\psi_i\}_{i = 1,N_{\psi}}$ such that $\forall t$, $
Q_t(s,a) = \sum_{i}{\alpha_{ti} \psi_i(s,a)} =  \langle \alpha_{t}, U^T \Phi(s,a) \rangle $ 
where $\psi_i$-s form a basis for the relevant low-dim subspace. 

Thus for each task $t$, we are trying to solve jointly the following optimization problem:
\begin{equation}
\arg \min_{f = \langle \alpha_t,U^T \Phi \rangle} {\left[\sum_{t}{\mathcal{L}_{\mathcal{D}_t}\left(f_t(s,a) , y_{t}^{(k)} \right)} + \Vert \alpha_t \Vert_{1}^2 \right] } 
\end{equation}

In \cite{Argyriou2008} this was shown to be equivalent to Eq. \ref{eq:MTL_Convex} and can be solved efficiently by an alternating minimization procedure:
\begin{equation}
\arg \min_{A,U} {\left[\sum_{t}{\mathcal{L}_{\mathcal{D}_t}\left(\langle \alpha_t, U^T \Phi(s,a) \rangle, y_{t}^{(k)} \right)} + \mathcal{H}(A)\right] }
\label{eq:MTL_Convex}
\end{equation}
where $A = [\alpha_1,\cdots, \alpha_t, \cdots, \alpha_T]$ and this is assumed to be sparse, take $\mathcal{H}(A) = \gamma ||A||_{2,1}^2$.

\subsection{Allowing for task specificity}
The above procedure can be used to construct very informative shared features -- as shown in \cite{Calandriello2014} and in our experimental section. However, in a lot of scenarios tasks can benefit from having a small and sparse set of features that represent the particularities of each individual task on top of a low-dimensional shared subspace.   This is definitely the case in many practical applications and had been observed in purely supervised settings as well -- it is simply too restricted to constrain all tasks to be using a single shared structure. Thus researchers have come up with various ways of incorporating task-specific components --- see \cite{Zhou2011}, \cite{Jalali2010},  \cite{Chen2012} and reference therein -- and showed that modelling these explicitly can improve both the learning (accuracy and speed) and interpretability of the resulting representations. In this work, we choose just one of these formulations, introduced in \cite{Ando2005}, where we learn a low-dimensional shared representation $U\Phi(s,a)$ as before, as well as a task specific vector $w_t$, on which we place a strong sparsity constraint to encourage common features to still be identified and shared. 
\begin{equation}
Q_{t, \theta}(s,a) = (w_t^T+v_t^T U) \Phi(s,a)
\end{equation}
Note that if $U$ is a zero matrix, then we will be treating the task as completely independent and on the other hand if $w_t$ is zero for all tasks, we recover the previous formulation. Furthermore, we can place an orthogonality condition on the set of shared features inferred, by enforcing $UU^T = I$. The resulting optimization problem has the form:
\begin{equation}
 \arg \min_{U, \lbrace v_t, w_t \rbrace_{t=1}^T } {\left[\sum_{t}{\mathcal{L}\left((w_t^T+v_t^T U)\Phi , \hat{Q}_{t} \right)} + \mathcal{H}(W)\right] }
\end{equation}
and can be solved by Alternating Structure Optimization (ASO)  -- see \cite{Ando2005}.

\section{Experiments}

We assess the performance and behaviour of our proposed model and learning procedures in the $4$-room navigation task \cite{SuttonPrecupSingh1999}. The state space $\mathcal{S}$ is described by all valid positions an agent might take -- any position in the grid, but the wall and the agent has access to four actions $\mathcal{A} = \{ \rightarrow, \leftarrow, \uparrow, \downarrow\}$. We consider a deterministic dynamics in those directions and all walls are considered elastic -- bumping into walls has no effect on your state. Tasks are specified as target locations in the environment the agent need to navigate to. These will be sampled at random from the valid states in the environment. We do not specify a starting state --  agents need to learn to navigate to the selected goal position from any part of the environment. When the agent transitions to the goal states, it collects a positive reward. No other reward signal is provided. 

Since all of the proposed methods can be run off-policy, thus decoupling the experience gathering and the learning, we sample a modest amount of experience up front for each of the considered tasks $\mathcal{D}_t$. This can be done, in principle, by any behaviour policy, but in all our data-gathering we employ uniformly random exploration. 

Once data is gathered or provided, we can proceed with the learning. All experiments were conducted under a restrictive sample budget $|\mathcal{D}_t| \in \{500, 750, 1000\}$.  Firstly, we would like to compare the proposed joint-representation learning with its single tasks counterparts of FQI and FPI to see what effects, if any, enforcing and learning a shared representation would have. We assess the quality of the inferred greedy policies by the amount of reward they are able to produce under random starts -- this is a proxy for the real value function $V^{\pi_t}$, where $\pi_t = \mathcal{G} \hat{Q}_t$.
\begin{equation*}
V^{\pi_t}_{emp} \approx \mathbb{E}_{s_0 \sim \mu}\left[\sum_{k=0}^{\infty} \gamma^k r_k \vert s_0, \pi_t\right] \approx V^{\pi_t} 
\end{equation*}

\begin{figure*}
	\centering
	\includegraphics[width = 0.9\textwidth, height = 0.3\textheight]{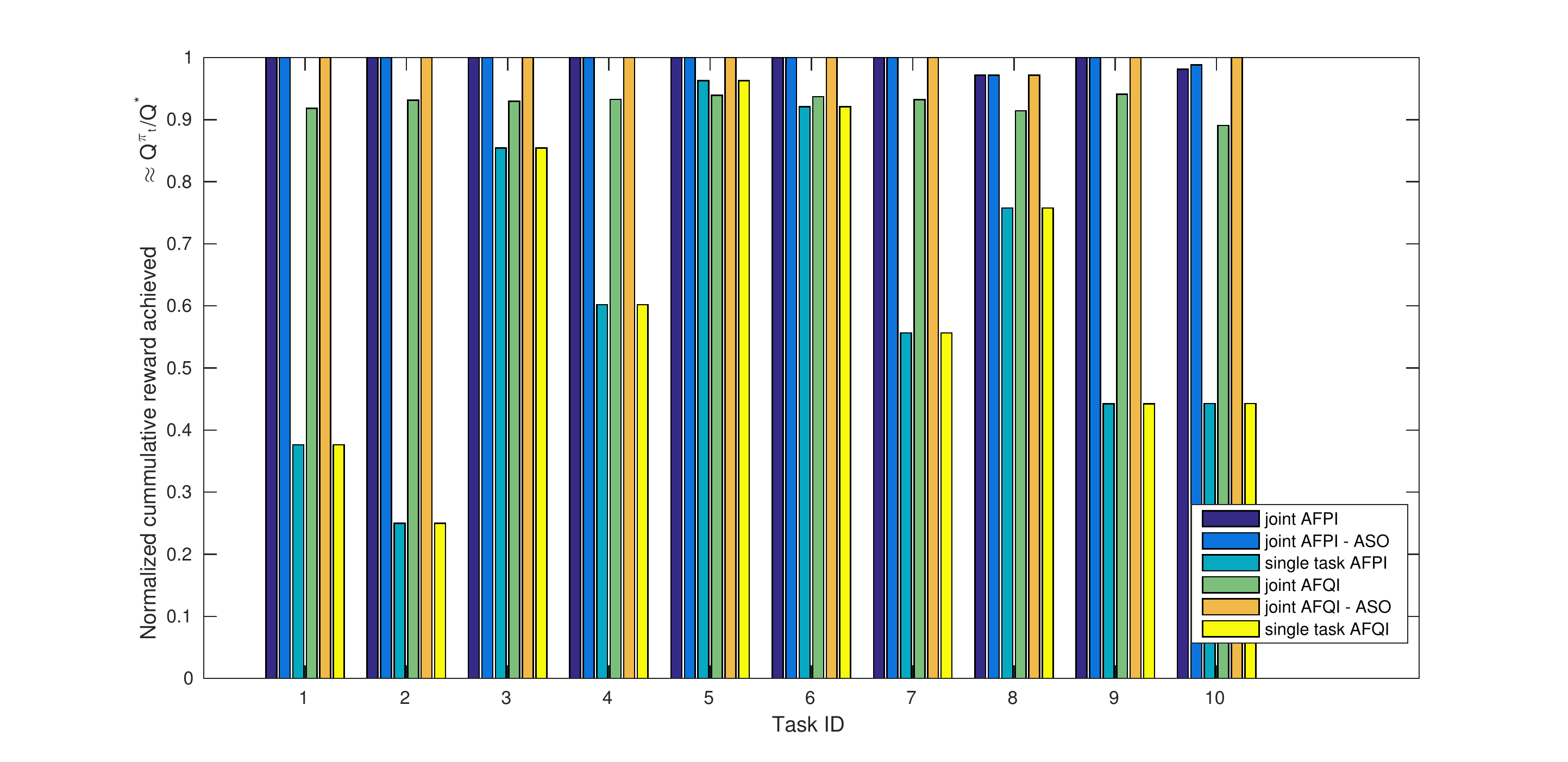}
	\caption{Quality of the inferred greedy policies, when trained individually and jointly on $30$ tasks, with a sample budget of $500$ samples/task. We show the average cumulative reward achieved by the agent from random initial positions -- the values are normalized with respect to the optimal cumulative reward achievable from the same starting positions $V^{\pi_t}_{emp}/V^{*}_{emp}$. We can see that in most cases, the single-task learning struggles under this sample regime, whereas the joint-learning methods are able to discover much better policies and even recover the optimal ones.}
\label{fig:policy_perfomance}
\end{figure*}
\begin{figure}
\begin{subfigure}
	\centering
	\includegraphics[width = 0.45\textwidth]{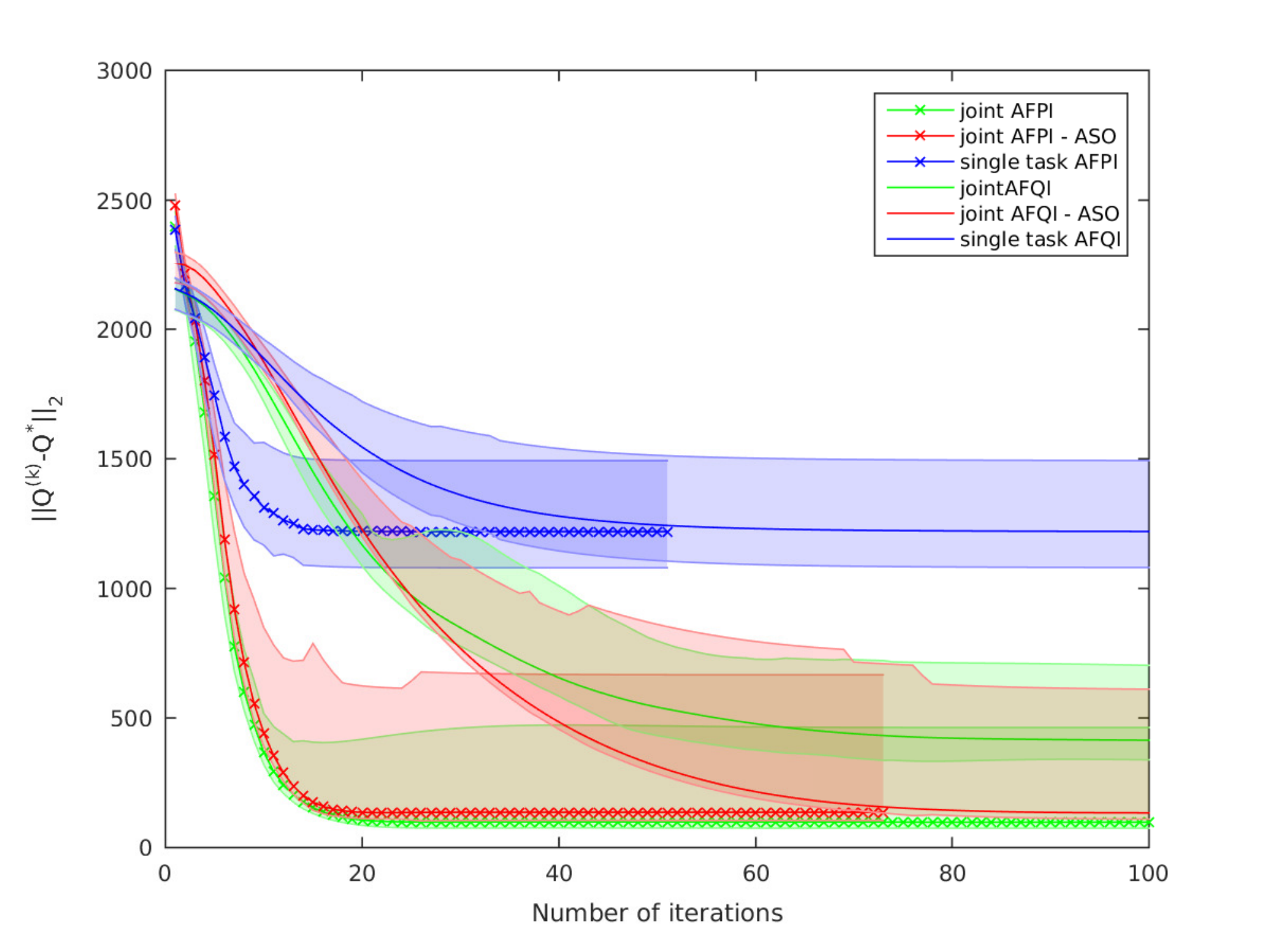}
\end{subfigure}
\begin{subfigure}
	\centering
	\includegraphics[width = 0.45\textwidth]{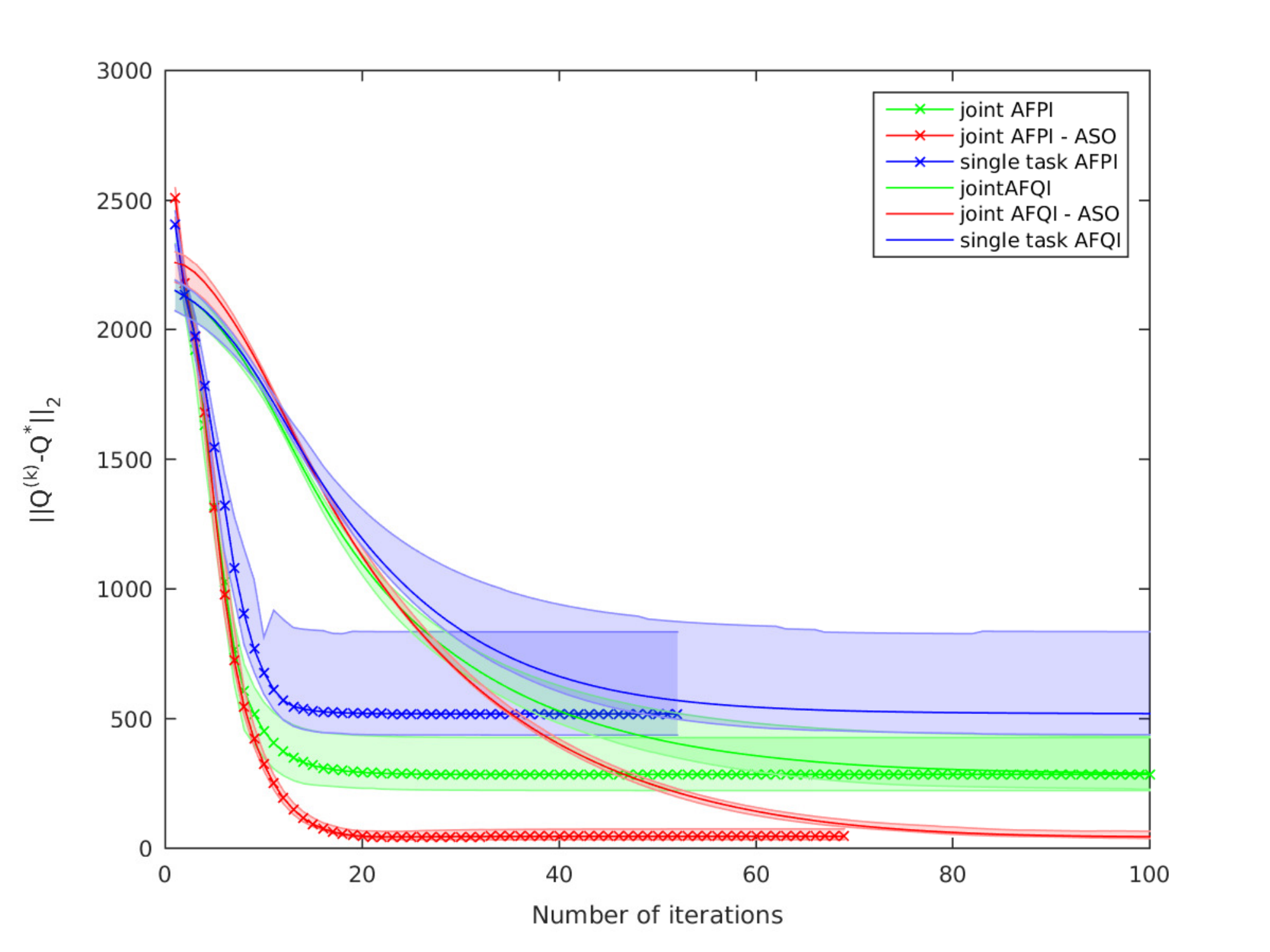}
\end{subfigure}
\begin{subfigure}
	\centering
	\includegraphics[width = 0.45\textwidth]{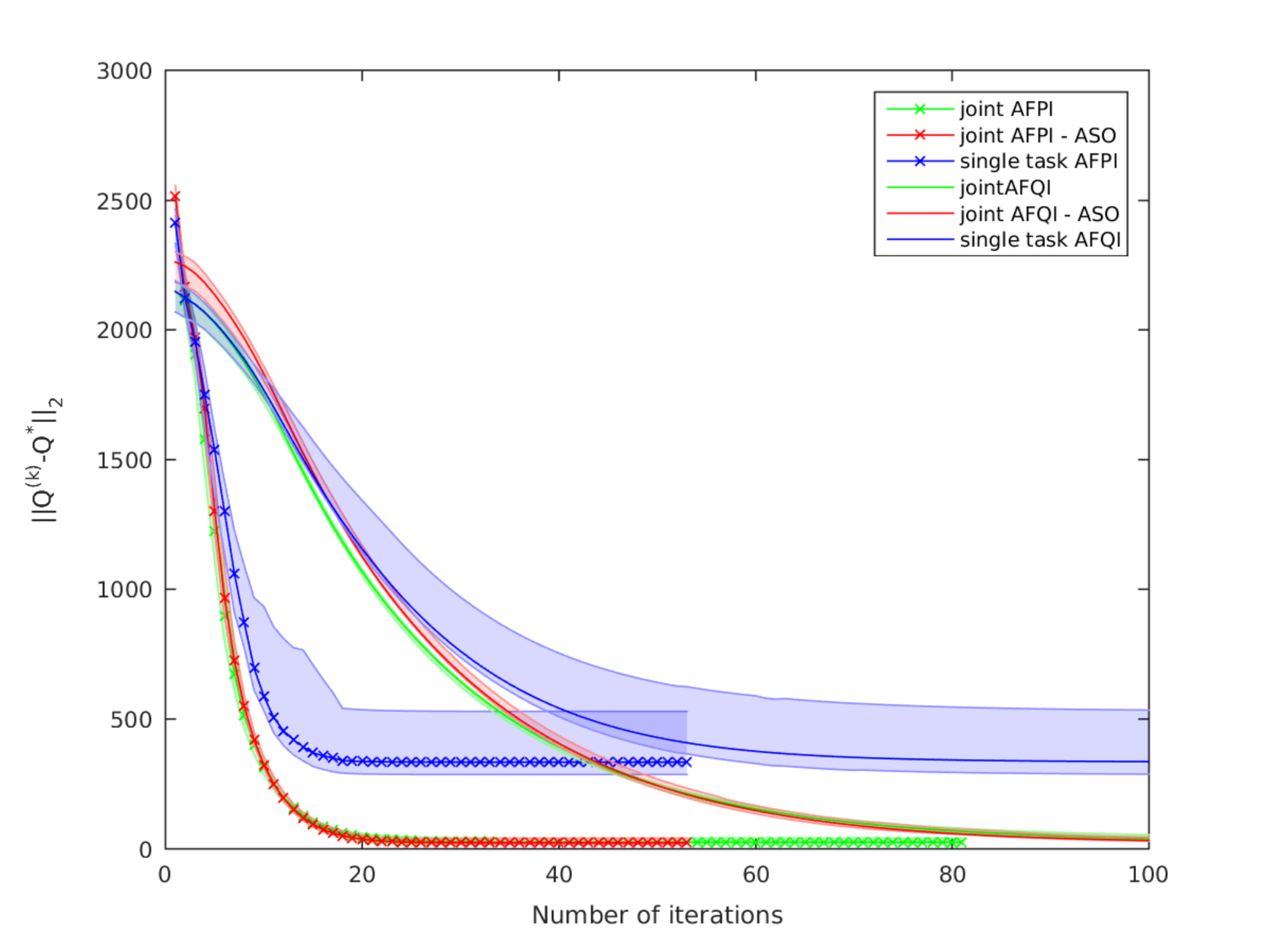}
\end{subfigure}
\caption{Convergence to optimal value function $Q^*$, as assessed by the Euclidean norm $||Q^*-Q^{(k)}||$ for the different sample complexities: 500 samples/task \textbf{(top)}, 750 samples/task \textbf{(middle)}, 1000 samples/task \textbf{(bottom)} for the different methods proposed. We report an average over $T=30$ tasks and shaded area corresponds to variances below and above the mean. Note that at $1000$ sample/task, all joint-representation learning algorithms obtain convergence to the true optimal value functions \textit{for all tasks}.  Also note that between the join-learning methods, the second method, allowing for task-specificity(red lines -- AFPI-ASO, AFQI-ASO) yields the better approximations $Q^{(k)}$. }
\label{fig:convergence_to_optimal}
\end{figure}
\begin{figure}
\begin{subfigure}
	\centering
	\includegraphics[width = 0.45\textwidth]{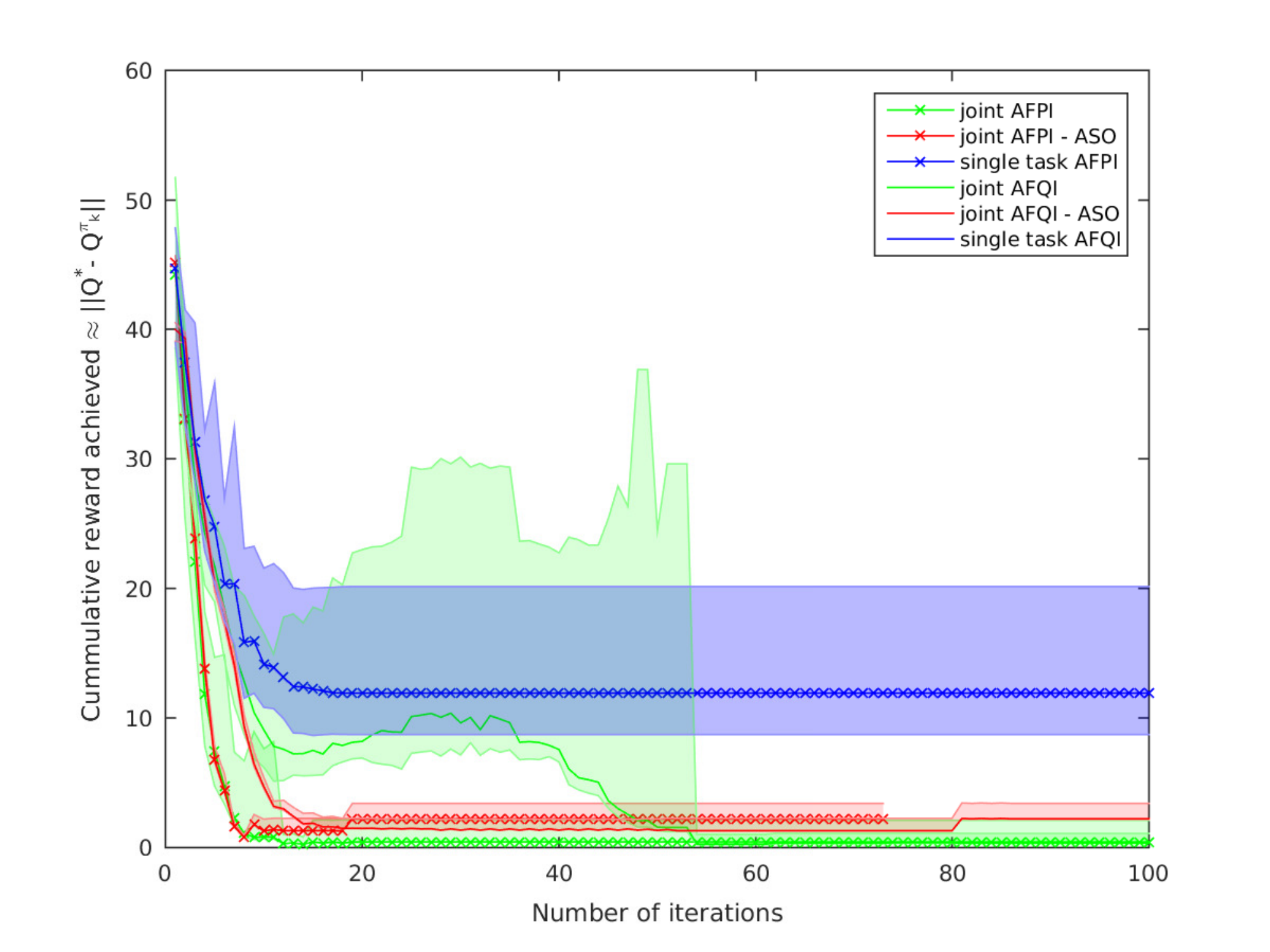}
\end{subfigure}
\begin{subfigure}
	\centering
	\includegraphics[width = 0.45\textwidth]{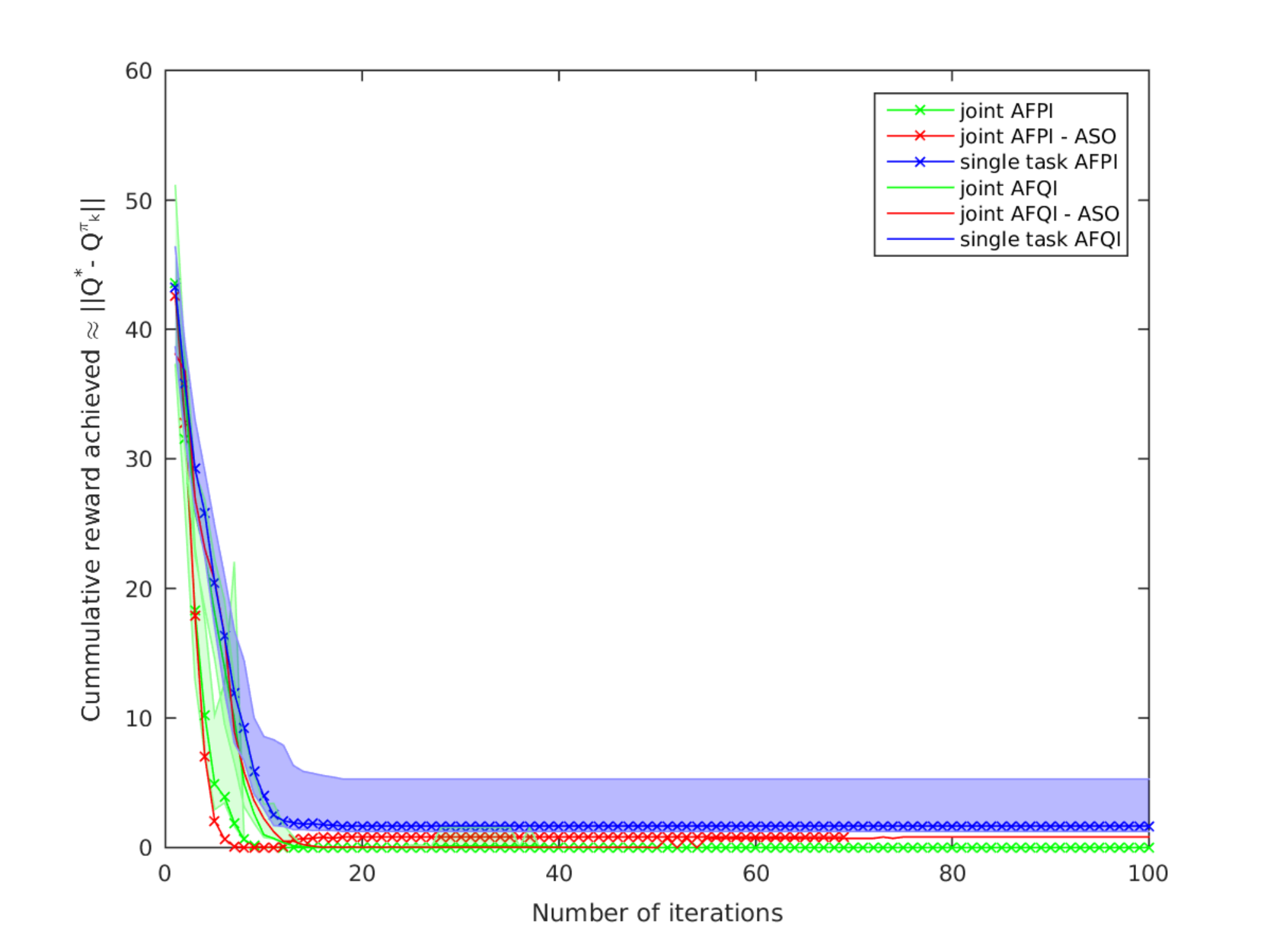}
\end{subfigure}
\begin{subfigure}
	\centering
	\includegraphics[width = 0.45\textwidth]{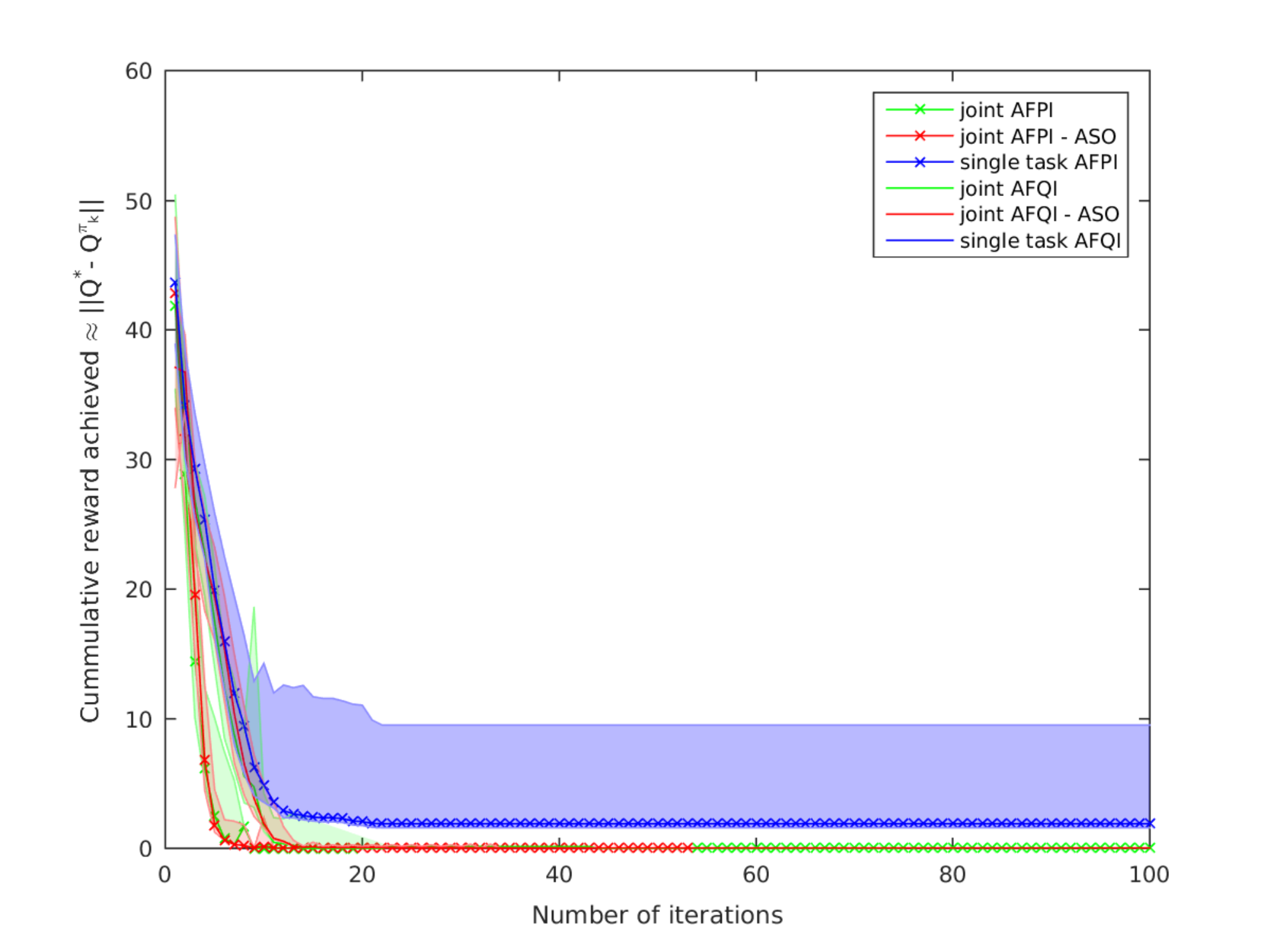}
\end{subfigure}
\caption{To evaluate the \textbf{quality of the policy learnt}, we produce an empirical estimate of $\frac{1}{T}\sum_t||V^*_t - V^{\pi_t}_t||/||V^{*}_t||$ for different sample complexities: 500 samples/task \textbf{(top)}, 750 samples/task \textbf{(middle)}, 1000 samples/task \textbf{(bottom)} for the different methods proposed. As seen from the convergence plot above, for $1000$ samples the multi-task methods will reliably recover the optimal value functions and implicitly the optimal policies, but so will the single-task methods in a lot of these tasks. At the same time for half that budget (500 samples), the multi-task learning is already able to recover the optimal policies, while the single-task methods \textit{converge} to a suboptimal value function (blue lines on the plot).}
\label{fig:creward_plots}
\end{figure}

Depending on the selection of starting states, the difficulty of the tasks and thus amount of reward achievable may vary. To ease interpretation, we report the normalized value of the above estimate, with respect to the optimal value function at the starting states. Example results for the first $10$ training tasks are displayed in Figure \ref{fig:policy_perfomance}. These were obtained by training on $30$ randomly sampled tasks on $500$ samples of experience per task. We can see that the joint-learning procedures manage to learn good policies, quite close to optimal ones that substantially outperform  the single-task learning. Please note that our proposed extension to allow task-specific features, in most cases, improves performance, even when considering a very small set of common features ($d_{shared} = 5$) - which also gives us a much faster convergence in the shared subspace. Indeed, this behaviour seems to be consistent to lower samples sizes, although it is worth mentioning that divergence does occur more often in these extrem conditions (very few samples) and regularization parameters that might ensure convergence \cite{Calandriello2014} provide a solution that is often worse than even the single task. Outside, those extreme cases, policy and value iteration methods perform very similarly and as we can see from Figure \ref{fig:convergence_to_optimal} -  \ref{fig:creward_plots}, that they tend to converge to the same solution.


To get a better idea of the average task performance we obtain and how that changes during training, we can look at the average distance between our estimate of the value functions at iteration $k$ and the optimal ones $Q^*_t$. For this small environment, these can be computed analytically. Results for $500$ and $1000$ samples budgets are displayed in Figure \ref{fig:convergence_to_optimal}. We observe quite a big difference between the single-task and multi-task procedures in terms of recovering the true optimal value functions. Convergence to a better MSE happens much faster and we get even asymptotic superior solution. Nevertheless, closeness to the optimal value functions in Euclidean space may not necessarily imply the same relation in policy space. A plot of the quality of policies as a function of value/policy iterations is available in Figure \ref{fig:creward_plots}. Here, we report the normalized average regret $\frac{1}{T}\sum_t||V^*_t - V^{\pi_t}_{emp,t}||/||V^{*}_t||$. We can see that the policies in general will converge much faster than the value functions, when comparing with the Q-value convergence in Figure \ref{fig:convergence_to_optimal}.
Please also note that the multi-task Fitted Policy-Iteration procedures inherit the same speedy convergence present in the single-task counterpart.

\subsection{Learnt shared representations}
\begin{figure}
\centering
	\includegraphics[width = 0.5\textwidth]{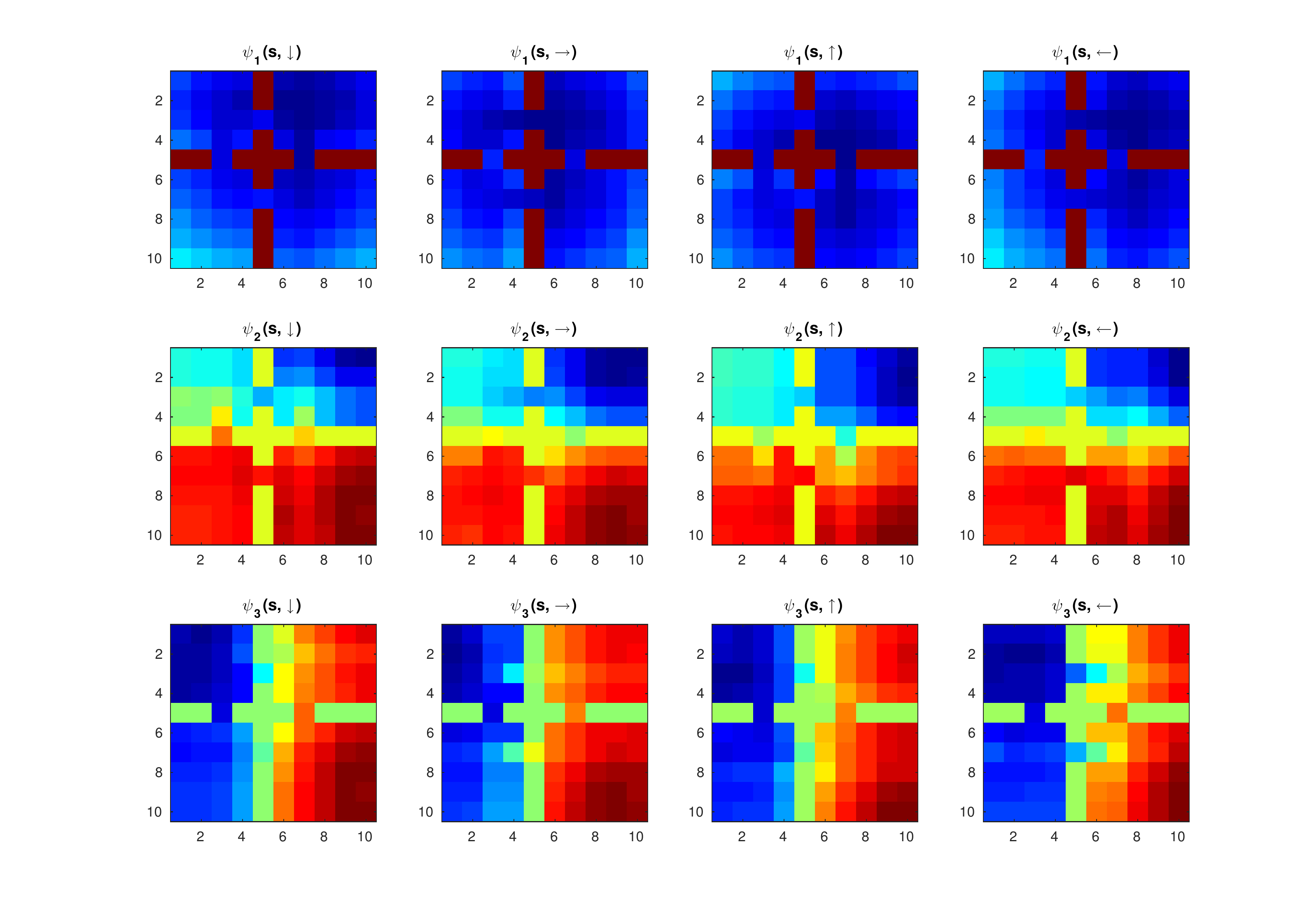}
	\caption{[To be read \textbf{row-wise}] The first three most relevant shared features $\psi_{1:3}(s,a)$ -- corresponding to the top three eigenvalues -- learnt via AFPI-MTFL under $30$ tasks randomly sampled in the four rooms.  Please note that these already enable the navigation between any pair of rooms.}
	\label{fig:features}
\end{figure}

\begin{figure}
\begin{subfigure}
\centering
	\includegraphics[width = 0.5\textwidth,height = 0.12\textwidth]{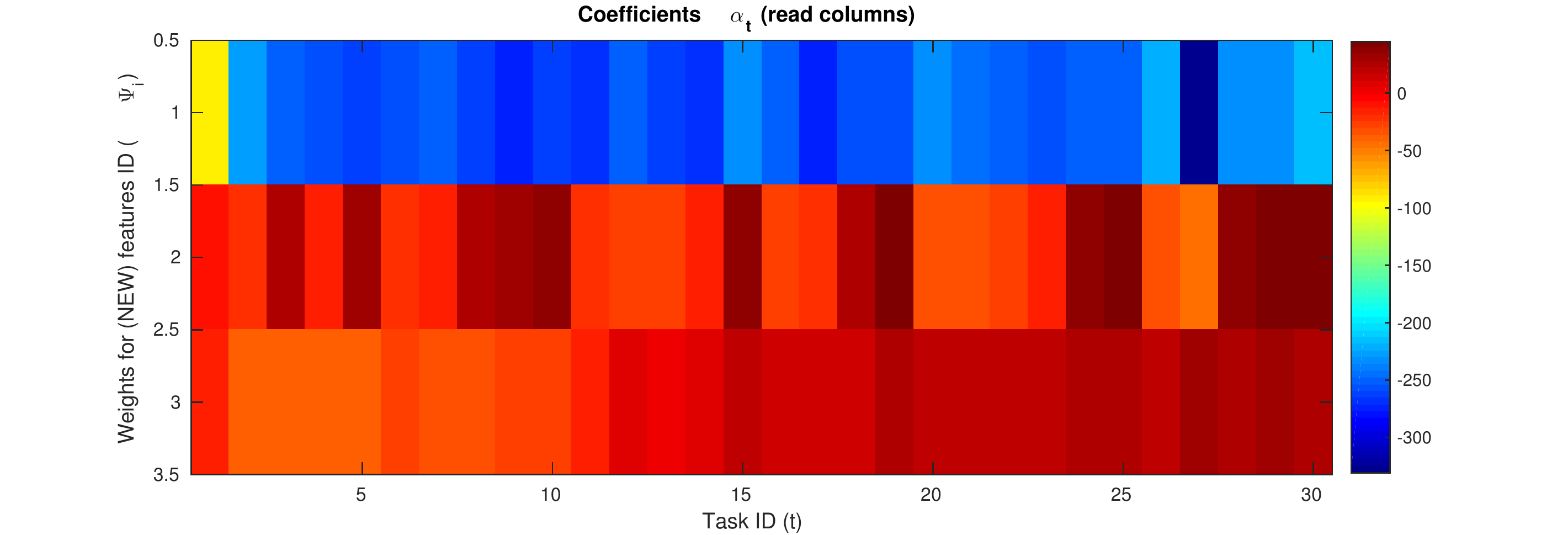}
\end{subfigure}
\begin{subfigure}
\centering
	\includegraphics[width = 0.5\textwidth,height = 0.08\textwidth]{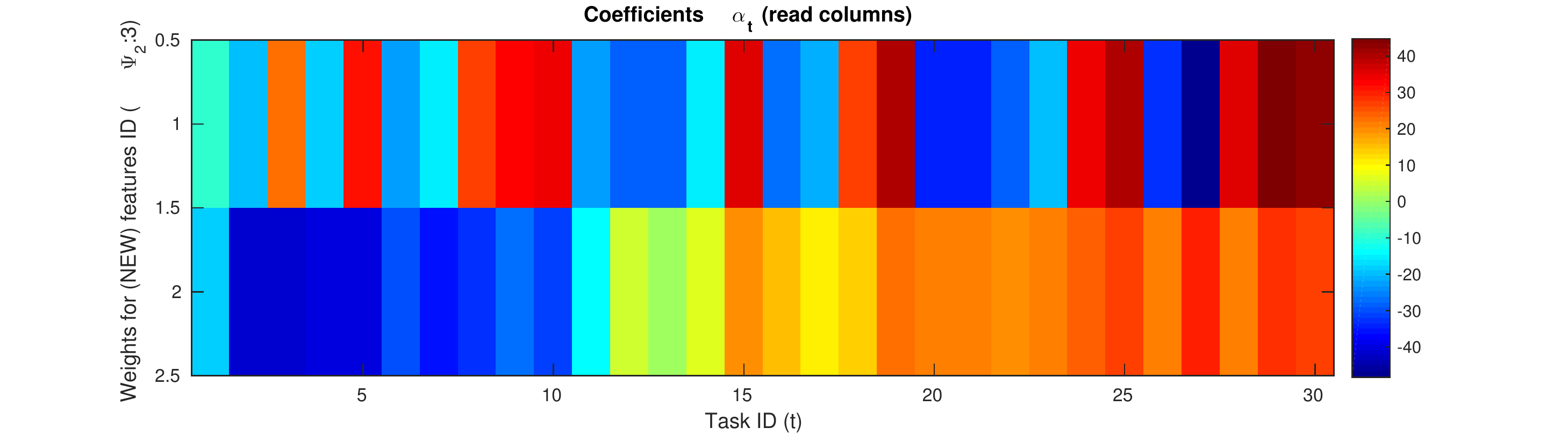}
\end{subfigure}
\caption{Weighting Coefficients $\alpha_{1:3,t}$ and $\alpha_{2:3,t}$ for the above three most prominent shared features. We can see from these values that the first feature clearly dominated in all tasks. \textbf{Bottom:} Rescaled version of $\alpha_{2:3,t}$  such that we can see the activation of the other two prominent features. Blue corresponds to negative activation and red to positive ones. Given the nature of the features one can readily read out, just by looking at the sign of the weight, which room that task's goal state is. For instance, if we look at second task: negative activation for both $\psi_2 \Rightarrow $ north-side of the environment and $\psi_3 \Rightarrow$ west-side of the environment. The goal $G_2$ is indeed located at position $(2,1)$ in the top-left room.}
\label{fig:weights}
\end{figure}

Probably the most interesting phenomenon encountered in learning these shared representations is the nature of the low dimensional representations inferred. We visualize the inferred set of shared features (Figure \ref{fig:features}) and their respective weights in the value-function (Figure \ref{fig:weights}). 

These were produced via MT-FQI with ASO, with the constraint that the shared subspace has at most $5$-dimensions. And even this seem to be too permissive, as we actually obtain strong activations only for the top 3 features inferred $\psi_{1:3}$ -- presented in Figure \ref{fig:features}. Thus the learnt representation is very low dimensional, but at the same time expressive enough to effectively approximate optimal value functions.
\subsection{Transferring knowledge to new tasks}

\begin{figure}
	\centering
	\includegraphics[width = 0.45\textwidth]{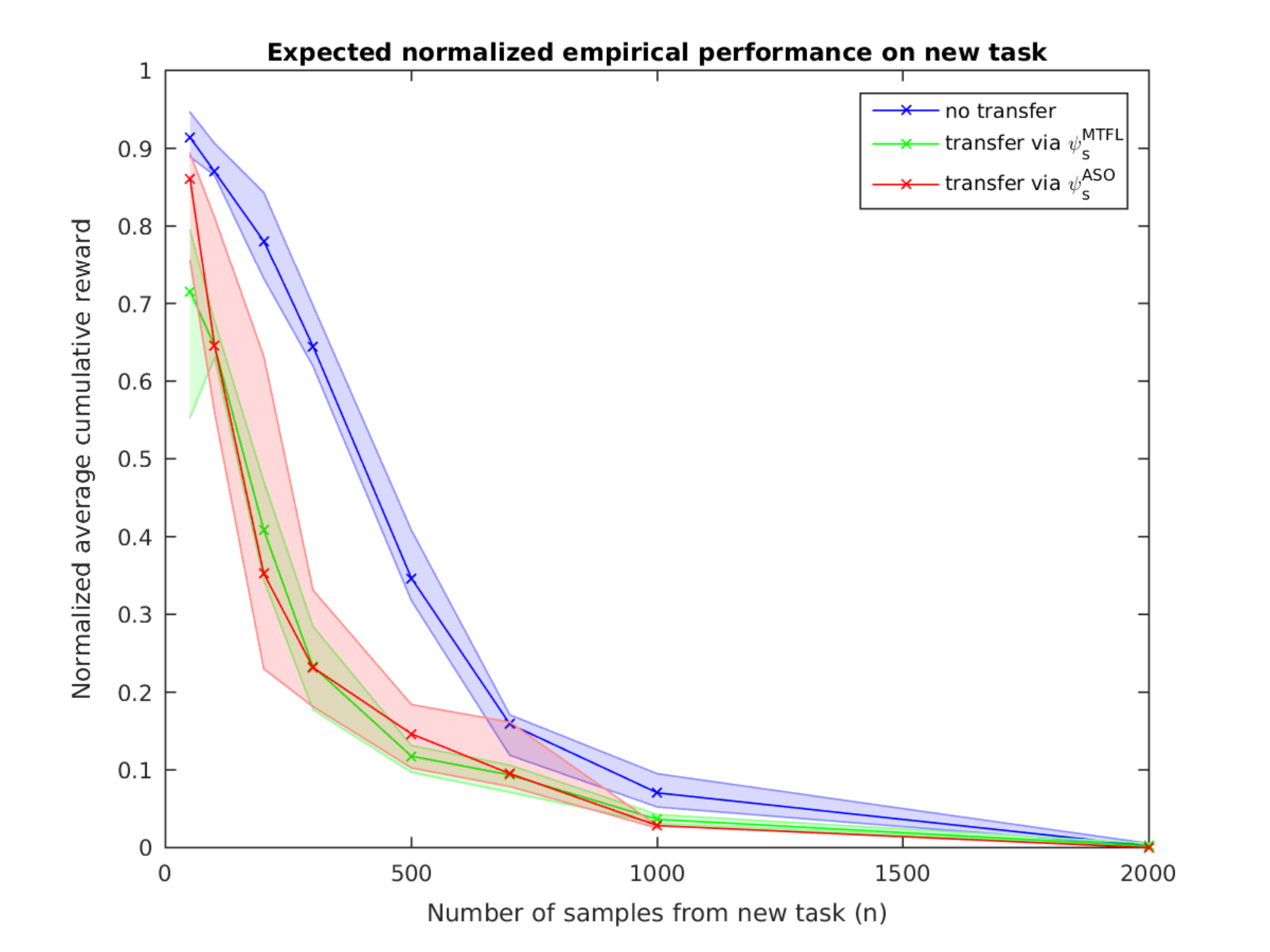}
	\caption{Average performance on a set of 10 new tasks with and without transfer of shared features, as assessed by the normalized average cumulative reward collected over 50 random starts in the environment. The value functions for the new tasks were produced by (single-task) FQI on original feature $\phi$ (no transfer), and respectively the augmented features space $[\psi_s^{MTFL}, \phi]$ and $[\psi_s^{ASO}, \phi]$ (transfer).}
	\label{fig:transfer}
\end{figure}

The learnt representations resemble option-like features  \cite{SuttonPrecupSingh1999} that essentially inform the agent, across tasks, how to navigate efficiently between rooms and negotiate the narrow hallways. 
These are indeed easily transferable 'skills' that can be use in learning a new task. We test this hypothesis by augmenting the representation for the new task, with this shared subspace. 
We investigate the benefits of having learnt a shared subspace over a set of training tasks in terms of transferring that knowledge when optimizing for a new task. We augment the feature space for the new task, with the learnt features $\psi_s$ and then we assess the effect this modification has on learning the new task. In Figure \ref{fig:transfer}, we present an empirical evaluation of the cumulative regret the agent will incur on the inferred (greedy) policy, when trained on the original representation $\phi$, versus the augmented representation $\{\psi_s, \phi\}$, after seeing a varying amount of samples $N = 50,200,300,500,700,1000, 2000$. We can see that the augmented representation is able to produce a good performance under smaller sample sizes. In general, the learning based on the transferred representation is able to produce a policy that is equivalent to the ones we could learn without transfer under twice as much data. This behaviour is consistent until convergence.

\subsection{Connection to Options}
As previously, the learnt shared representation seems account for the general topology and dynamics of the environment in the value functions. They nicely partition the environment into relevant regions to facilitate the global navigation to a local neighbourhood of the goal. Some of those features are characteristic of options \cite{SuttonPrecupSingh1999}, skills \cite{KonidarisBarto2007a}, macro-actions literature \cite{Dietterich2000} and are hve the potential to drastically improve the efficiency and scalability of RL methods \cite{BartoMahadevan2003}, \cite{Hengst2002}. In the following we would like to investigate this connection further.

Following the formulation in \cite{SuttonPrecupSingh1999}, an option $o = \langle \mathcal{I}, \mu, \beta \rangle$, is a generalization of primitive actions $a \in \mathcal{A}$ to a temporally extended course of action. $\mathcal{I}$ is the initiation set $\mathcal{I} \subseteq \mathcal{S}$ from which the option is available, $\mu$ is the policy we are going to follow once the options is triggered and $\beta:\mathcal{S} \rightarrow [0,1]$ is the probability of termination. In this case, the value function take the form:
$$
Q^{\pi_t}_t(s,o) = \mathbb{E}\left[ \underbrace{\sum_{k=0}^{K} \gamma^k r_{t,k}}_{r_t(s,o)} + \gamma^{K+1} \sum_{s'} P(s'|s,o) V^\pi_t(s') \right] 
$$
where $s'$ is in the termination state of options $o$.
We denote $P^o_{ss'} = \sum_{k}^{\infty}{p_o(s',k)\gamma^k}$, where $p_o(s',k)$ is the probability that options $o$ will terminate in state $s'$ after exactly $k$ steps. Note that this term accounts for the transition dynamics, the policy of the option and its termination criteria, all of which, for us, are task-invariant.
Moreover note that for us, $r_t(s,o)$ is generally $0$ unless the option happens to hit the goal. Thus the above equation, simplifies to:
$$
Q^{\pi_t}_t(s,o) = \sum_{s'} \left[ \underbrace{P^o_{ss'}}_{ \text{ task independent} (t, \pi_t)} \cdot \underbrace{ V^{\pi_t}(s') }_{\text{task} \text{ dependent} (t, \pi_t)} \right] 
$$
This is a linear combination between the option transition models $P^o_{ss'} = \phi^{\mu_o}(s, s')$ in the termination set (subgoals of the option) -- which is independent of task $t$ and $\pi_t$, it only depends on $\mu_o$ -- weighted by the value function of the termination states for each of the tasks -- which incorporates the dependency on task and the individual policy employed after the option has terminated. This is very similar to the parametrization we assumed in Eq. \ref{eq:linear_parametrization_value_function}. This suggest that the learnt representation is able to capture and represent efficiently some option-like transition models without specifying any subgoals, policies nor initial states. 
We hypothesize that \textit{the learnt shared space is actually a compressed basis for these option-transition models}. In order to test this hypothesis, we consider an intuitive set of options $\mathcal{O}$ (like navigation to a particular room) and test if this learnt basis can span $P^{o}_{ss'}$ for some option $o \in \mathcal{O}$ and can successfully represent an option-policy.

\begin{figure}[ht]
\begin{subfigure}
	\centering
	\includegraphics[width = 0.23\textwidth ]{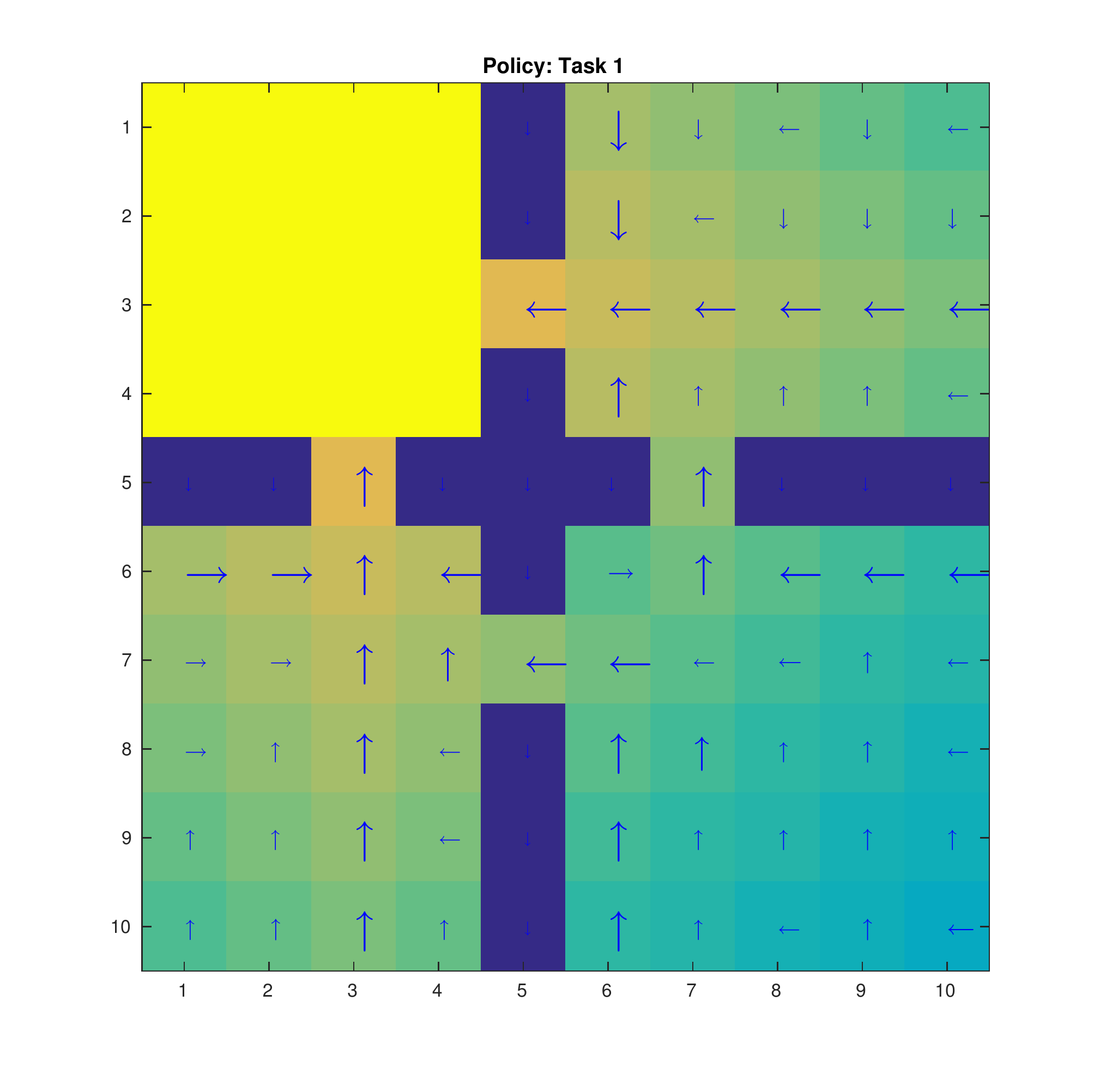}
\end{subfigure}
\begin{subfigure}
	\centering
	\includegraphics[width = 0.23\textwidth]{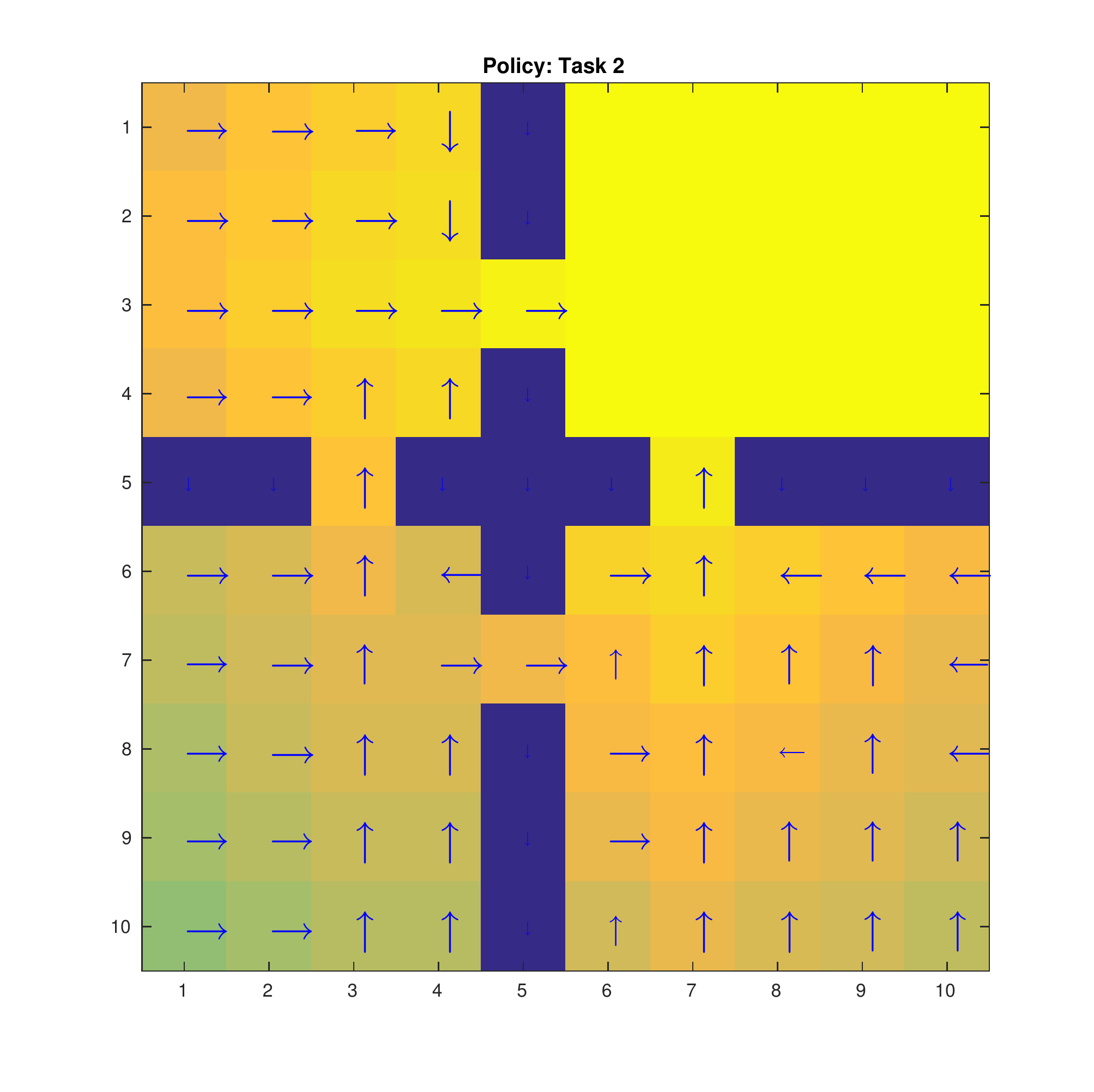}
\end{subfigure}

\begin{subfigure}
	\centering
	\includegraphics[width = 0.23\textwidth]{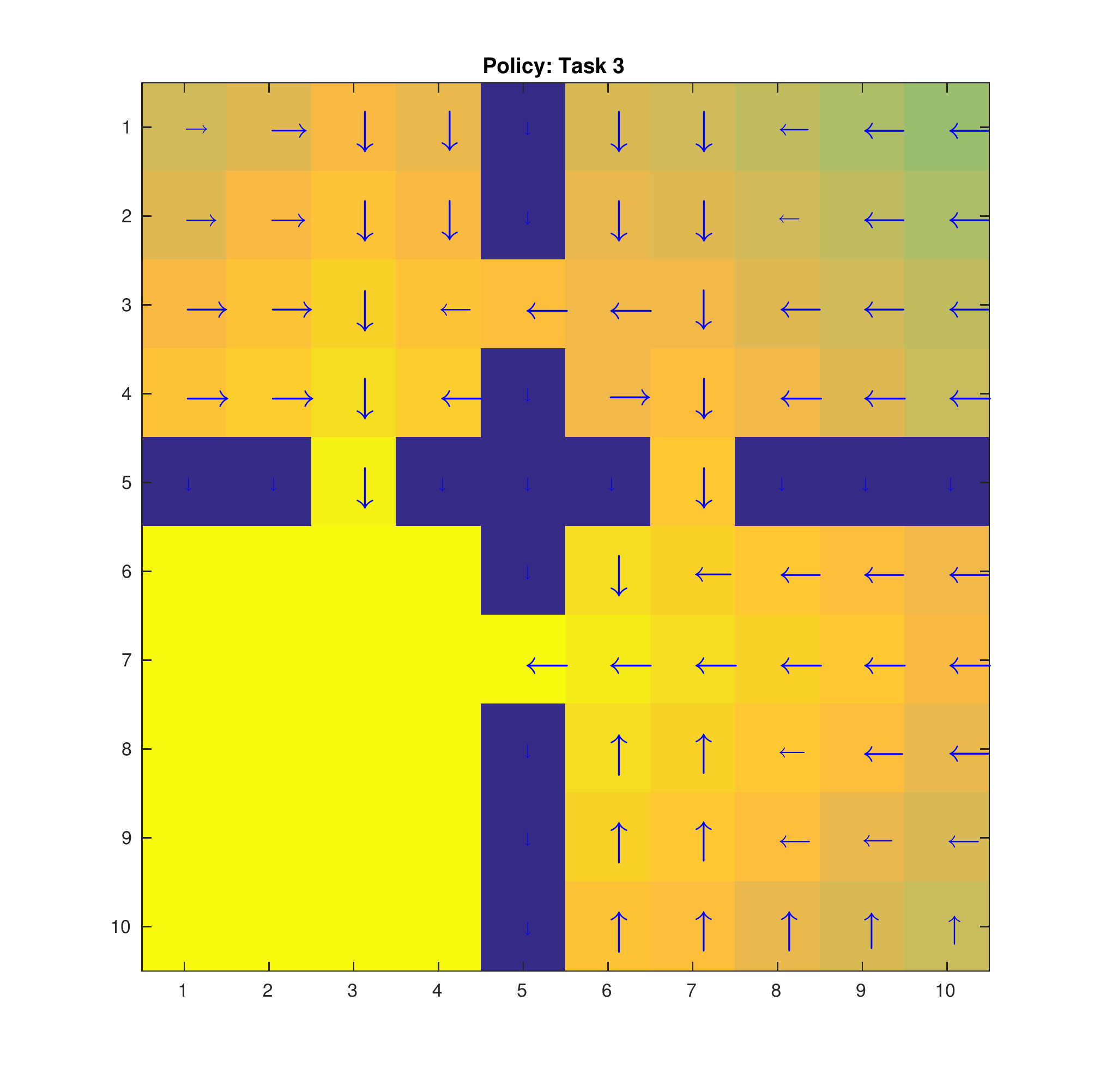}
\end{subfigure}
\begin{subfigure}
	\centering
	\includegraphics[width = 0.23\textwidth]{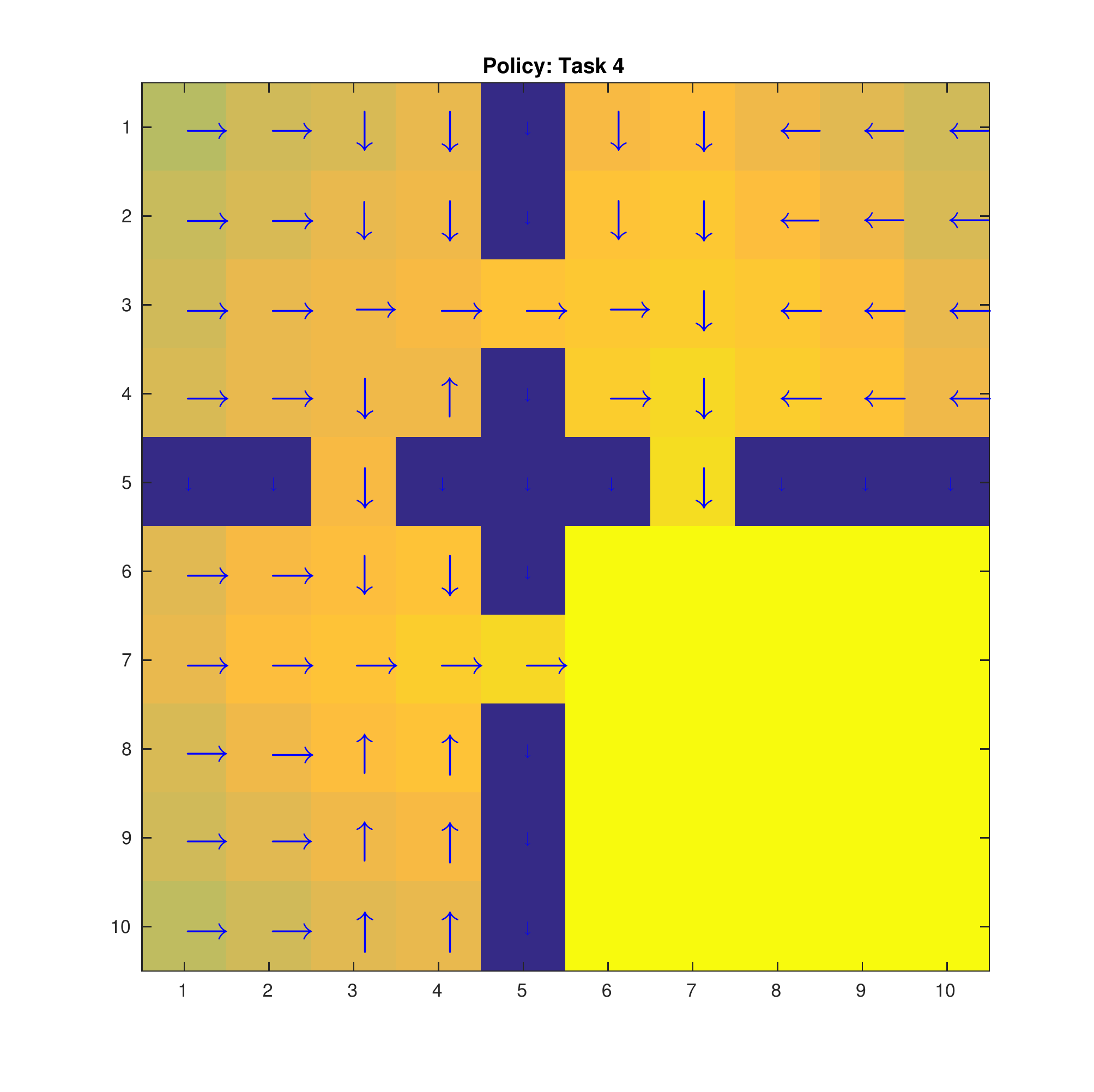}
\end{subfigure}
\caption{Learned greedy policies (as indicated by the arrows) and value functions ($V(s) = \max_a(Q(s,a))$ ) enabling navigation to any of the four rooms, based only on the share feature subspace discovered in the multi-task value function learning of $30$ goals randomly sampled in the environment. The value functions were learnt using (single-task) FQI on top of features $\psi^{ASO}_s$ and required 200 samples to recover option-like policies that enable the agent to reach the desired room.}
\label{fig:options_rooms}
\end{figure}

We define an option $o_1$ to be navigating to a specific room, say room $1$ (NW). The initialization set $\mathcal{I}_{o_1}$ is the set of states outside the room and termination set is any state in the desired room. We also can define an MDP that maintains the same transition dynamics, state and action space, but now the reward signal is zero outside the target room and a constant positive reward in any of the desired termination states $s' \in$ room $1$. Note that the value function corresponding to this newly defined semi-MDP \cite{SuttonPrecupSingh1999} is given by: $Q^{\pi}(s,o) = \sum_{s'} {P^o_{ss'}}$, as $V^{\pi}(s') = const$ \footnote{This is actually true, only under a mild assumption that the agent under $\pi(s')$ will not leave the room, which is where all the reward is.}.
In this semi-MDP we run FQI and indeed see that we are able to construct a value function, based solely on the learnt $5$-dim feature space $\psi_s$, that successfully completes the specified task. Results for all such navigation options are available in Figure \ref{fig:options_rooms}.

Please note that the above defined options are quite extended ones. Simpler ones would include making your way outside a particular room -- these are along the lines of the options defined in \cite{SuttonPrecupSingh1999} and \cite{StollePrecup2002} -- and these can be easily recovered as well. Actually for these simpler options we require very few samples to obtain the desired behaviour (10-30 samples), although they might not be optimal -- please consult Supplementary material for details. 
The fact that we are able to express a whole variety of such intuitively defined options -- much more than the dimensionality of the common subspace on which we are building on -- is a clear indication of the expressiveness of this shared representation and its potential transferability in aiding learning of new tasks within the same environment.

\section{Related work}
There is a good collection of methods that tackle various aspect of multi-task reinforcement learning \cite{Lazaric2012} and \cite{Taylor2009}. As with our approach, these methods try to learn jointly either value functions or policies over a set of tasks \cite{LazaricGhavamzadeh2010}, \cite{DimitrakakisRothkopf2011} but under different structure and environment assumptions. A more recent study in \cite{KonidarisScheidwasserBarto2012}, also employs the idea of a shared feature space, but both the learning procedure and the proposed way of transferring this tuned knowledge is very different from ours.  
The main novel idea this work introduces is modelling explicitly a shared abstraction of the state-action space that can be refined throughout the learning process while optimizing for the value functions. The ability to change the representation throughout the learning process to model the improving set of policies is crucial. This is the only way option-like features could emerges -- these already incorporate both the transition model and good policies that generalize over tasks, as shown in the previous section. One of the methods investigated in \cite{Calandriello2014} in a study on sparsity in multi-task, is very closely related to our learning procedure and this work can be seen as a generalization of that method, although the focus and model assumptions are quite different. Perhaps the most relevant prior work that shares our vision and some of the modelling assumption is the approach in \cite{SchaulHorganGregorEtAl2015} which model a shared state-representation between goals and assume a linear factorization between this state embbeding and task/goal embbeding. 
 
\section{Conclusion and Future work}
In this work, we investigated the problem of representation learning in multi-task/multi-goal reinforcement learning. We introduced the multi-task RL paradigm and  showed how two of the most popular classes of planning algorithms, fitted Q-Iteration and approximate Policy Iteration, can be extended to learn from multiple tasks jointly. 
Focusing on linear parametrization of the $Q$-function, we showed at least two ways in which one can harness the power of well-established multi-task learning and transfer algorithms developed in supervised settings and apply them to inferring a joint structure over optimal value functions, and implicitly over policies.
As argued before and shown in these preliminary experiments, RL can benefit a lot from integrating joint treatment of goals and exploiting commonality between tasks. This ought to lead to more efficient learning and better generalization. Although these are very encouraging results, this paradigm does need more investigation to assess convergence behaviour, scalability to more complex tasks, employing other multi-task learning or representation learning procedures and we hope this work will serve as staring point.


\bibliographystyle{icml2015}
\bibliography{bibliography}

\begin{thebibliography}{26}
\providecommand{\natexlab}[1]{#1}
\providecommand{\url}[1]{\texttt{#1}}
\expandafter\ifx\csname urlstyle\endcsname\relax
  \providecommand{\doi}[1]{doi: #1}\else
  \providecommand{\doi}{doi: \begingroup \urlstyle{rm}\Url}\fi

\bibitem[Ando \& Zhang(2005)Ando and Zhang]{Ando2005}
Ando, Rie~Kubota and Zhang, Tong.
\newblock A framework for learning predictive structures from multiple tasks
  and unlabeled data.
\newblock \emph{The Journal of Machine Learning Research}, 6:\penalty0
  1817--1853, 2005.

\bibitem[Antos et~al.(2007)Antos, Szepesv{\'a}ri, and
  Munos]{AntosSzepesvariMunos2007}
Antos, Andr{\'a}s, Szepesv{\'a}ri, Csaba, and Munos, R{\'e}mi.
\newblock Value-iteration based fitted policy iteration: learning with a single
  trajectory.
\newblock In \emph{Approximate Dynamic Programming and Reinforcement Learning,
  2007. ADPRL 2007. IEEE International Symposium on}, pp.\  330--337. IEEE,
  2007.

\bibitem[Argyriou et~al.(2008)Argyriou, Evgeniou, and Pontil]{Argyriou2008}
Argyriou, Andreas, Evgeniou, Theodoros, and Pontil, Massimiliano.
\newblock Convex multi-task feature learning.
\newblock \emph{Machine Learning}, 73\penalty0 (3):\penalty0 243--272, 2008.

\bibitem[Barto \& Mahadevan(2003)Barto and Mahadevan]{BartoMahadevan2003}
Barto, Andrew~G and Mahadevan, Sridhar.
\newblock Recent advances in hierarchical reinforcement learning.
\newblock \emph{Discrete Event Dynamic Systems}, 13\penalty0 (4):\penalty0
  341--379, 2003.

\bibitem[Bengio(2009)]{Bengio2009a}
Bengio, Yoshua.
\newblock Learning deep architectures for ai.
\newblock \emph{Foundations and trends{\textregistered} in Machine Learning},
  2\penalty0 (1):\penalty0 1--127, 2009.

\bibitem[Calandriello et~al.(2014)Calandriello, Lazaric, and
  Restelli]{Calandriello2014}
Calandriello, Daniele, Lazaric, Alessandro, and Restelli, Marcello.
\newblock Sparse multi-task reinforcement learning.
\newblock In \emph{Advances in Neural Information Processing Systems}, pp.\
  819--827, 2014.

\bibitem[Chen et~al.(2012)Chen, Liu, and Ye]{Chen2012}
Chen, Jianhui, Liu, Ji, and Ye, Jieping.
\newblock Learning incoherent sparse and low-rank patterns from multiple tasks.
\newblock \emph{ACM Transactions on Knowledge Discovery from Data (TKDD)},
  5\penalty0 (4):\penalty0 22, 2012.

\bibitem[Dietterich(2000)]{Dietterich2000}
Dietterich, Thomas~G.
\newblock Hierarchical reinforcement learning with the maxq value function
  decomposition.
\newblock \emph{J. Artif. Intell. Res.(JAIR)}, 13:\penalty0 227--303, 2000.

\bibitem[Dimitrakakis \& Rothkopf(2011)Dimitrakakis and
  Rothkopf]{DimitrakakisRothkopf2011}
Dimitrakakis, Christos and Rothkopf, Constantin~A.
\newblock Bayesian multitask inverse reinforcement learning.
\newblock In \emph{Recent Advances in Reinforcement Learning}, pp.\  273--284.
  Springer, 2011.

\bibitem[Ernst et~al.(2005)Ernst, Geurts, and Wehenkel]{Ernst2005}
Ernst, Damien, Geurts, Pierre, and Wehenkel, Louis.
\newblock Tree-based batch mode reinforcement learning.
\newblock In \emph{Journal of Machine Learning Research}, pp.\  503--556, 2005.

\bibitem[Hengst(2002)]{Hengst2002}
Hengst, Bernhard.
\newblock Discovering hierarchy in reinforcement learning with hexq.
\newblock In \emph{ICML}, volume~2, pp.\  243--250, 2002.

\bibitem[Jalali et~al.(2010)Jalali, Sanghavi, Ruan, and Ravikumar]{Jalali2010}
Jalali, Ali, Sanghavi, Sujay, Ruan, Chao, and Ravikumar, Pradeep~K.
\newblock A dirty model for multi-task learning.
\newblock In \emph{Advances in Neural Information Processing Systems}, pp.\
  964--972, 2010.

\bibitem[Konidaris \& Barto(2007)Konidaris and Barto]{KonidarisBarto2007a}
Konidaris, George and Barto, Andrew~G.
\newblock Building portable options: Skill transfer in reinforcement learning.
\newblock In \emph{IJCAI}, volume~7, pp.\  895--900, 2007.

\bibitem[Konidaris et~al.(2012)Konidaris, Scheidwasser, and
  Barto]{KonidarisScheidwasserBarto2012}
Konidaris, George, Scheidwasser, Ilya, and Barto, Andrew~G.
\newblock Transfer in reinforcement learning via shared features.
\newblock \emph{The Journal of Machine Learning Research}, 13\penalty0
  (1):\penalty0 1333--1371, 2012.

\bibitem[Lazaric(2012)]{Lazaric2012}
Lazaric, Alessandro.
\newblock Transfer in reinforcement learning: a framework and a survey.
\newblock In \emph{Reinforcement Learning}, pp.\  143--173. Springer, 2012.

\bibitem[Lazaric \& Ghavamzadeh(2010)Lazaric and
  Ghavamzadeh]{LazaricGhavamzadeh2010}
Lazaric, Alessandro and Ghavamzadeh, Mohammad.
\newblock Bayesian multi-task reinforcement learning.
\newblock In \emph{ICML-27th International Conference on Machine Learning},
  pp.\  599--606. Omnipress, 2010.

\bibitem[Mnih et~al.(2015)Mnih, Kavukcuoglu, Silver, Rusu, Veness, Bellemare,
  Graves, Riedmiller, Fidjeland, Ostrovski, et~al.]{Mnih2015}
Mnih, Volodymyr, Kavukcuoglu, Koray, Silver, David, Rusu, Andrei~A, Veness,
  Joel, Bellemare, Marc~G, Graves, Alex, Riedmiller, Martin, Fidjeland,
  Andreas~K, Ostrovski, Georg, et~al.
\newblock Human-level control through deep reinforcement learning.
\newblock \emph{Nature}, 518\penalty0 (7540):\penalty0 529--533, 2015.

\bibitem[Schaul et~al.(2015)Schaul, Horgan, Gregor, and
  Silver]{SchaulHorganGregorEtAl2015}
Schaul, Tom, Horgan, Daniel, Gregor, Karol, and Silver, David.
\newblock Universal value function approximators.
\newblock In \emph{Proceedings of the 32nd International Conference on Machine
  Learning (ICML-15)}, pp.\  1312--1320, 2015.

\bibitem[Silver et~al.(2016)Silver, Huang, Maddison, Guez, Sifre, van~den
  Driessche, Schrittwieser, Antonoglou, Panneershelvam, Lanctot,
  et~al.]{SilverHuangMaddisonEtAl2016}
Silver, David, Huang, Aja, Maddison, Chris~J, Guez, Arthur, Sifre, Laurent,
  van~den Driessche, George, Schrittwieser, Julian, Antonoglou, Ioannis,
  Panneershelvam, Veda, Lanctot, Marc, et~al.
\newblock Mastering the game of go with deep neural networks and tree search.
\newblock \emph{Nature}, 529\penalty0 (7587):\penalty0 484--489, 2016.

\bibitem[Stolle \& Precup(2002)Stolle and Precup]{StollePrecup2002}
Stolle, Martin and Precup, Doina.
\newblock Learning options in reinforcement learning.
\newblock In \emph{SARA}, pp.\  212--223. Springer, 2002.

\bibitem[Sutton \& Barto(1998)Sutton and Barto]{Sutton1998}
Sutton, Richard~S and Barto, Andrew~G.
\newblock \emph{Reinforcement learning: An introduction}, volume~1.
\newblock MIT press Cambridge, 1998.

\bibitem[Sutton et~al.(1999)Sutton, Precup, and Singh]{SuttonPrecupSingh1999}
Sutton, Richard~S, Precup, Doina, and Singh, Satinder.
\newblock Between {MDP}s and semi-{MDP}s: A framework for temporal abstraction
  in reinforcement learning.
\newblock \emph{Artificial intelligence}, 112\penalty0 (1):\penalty0 181--211,
  1999.

\bibitem[Sutton et~al.(2011)Sutton, Modayil, Delp, Degris, Pilarski, White, and
  Precup]{SuttonModayilDelpEtAl2011}
Sutton, Richard~S, Modayil, Joseph, Delp, Michael, Degris, Thomas, Pilarski,
  Patrick~M, White, Adam, and Precup, Doina.
\newblock Horde: A scalable real-time architecture for learning knowledge from
  unsupervised sensorimotor interaction.
\newblock In \emph{The 10th International Conference on Autonomous Agents and
  Multiagent Systems-Volume 2}, pp.\  761--768. International Foundation for
  Autonomous Agents and Multiagent Systems, 2011.

\bibitem[Taylor \& Stone(2009{\natexlab{a}})Taylor and Stone]{Taylor2009}
Taylor, Matthew~E and Stone, Peter.
\newblock Transfer learning for reinforcement learning domains: A survey.
\newblock \emph{The Journal of Machine Learning Research}, 10:\penalty0
  1633--1685, 2009{\natexlab{a}}.

\bibitem[Taylor \& Stone(2009{\natexlab{b}})Taylor and Stone]{TaylorStone2009}
Taylor, Matthew~E and Stone, Peter.
\newblock Transfer learning for reinforcement learning domains: A survey.
\newblock \emph{The Journal of Machine Learning Research}, 10:\penalty0
  1633--1685, 2009{\natexlab{b}}.

\bibitem[Zhou et~al.(2011)Zhou, Chen, and Ye]{Zhou2011}
Zhou, Jiayu, Chen, Jianhui, and Ye, Jieping.
\newblock Clustered multi-task learning via alternating structure optimization.
\newblock In \emph{Advances in neural information processing systems}, pp.\
  702--710, 2011.

\end{thebibliography}

\newpage
\onecolumn

\begin{figure}[ht]
\begin{subfigure}
	\centering
	\includegraphics[width = 0.2\textwidth]{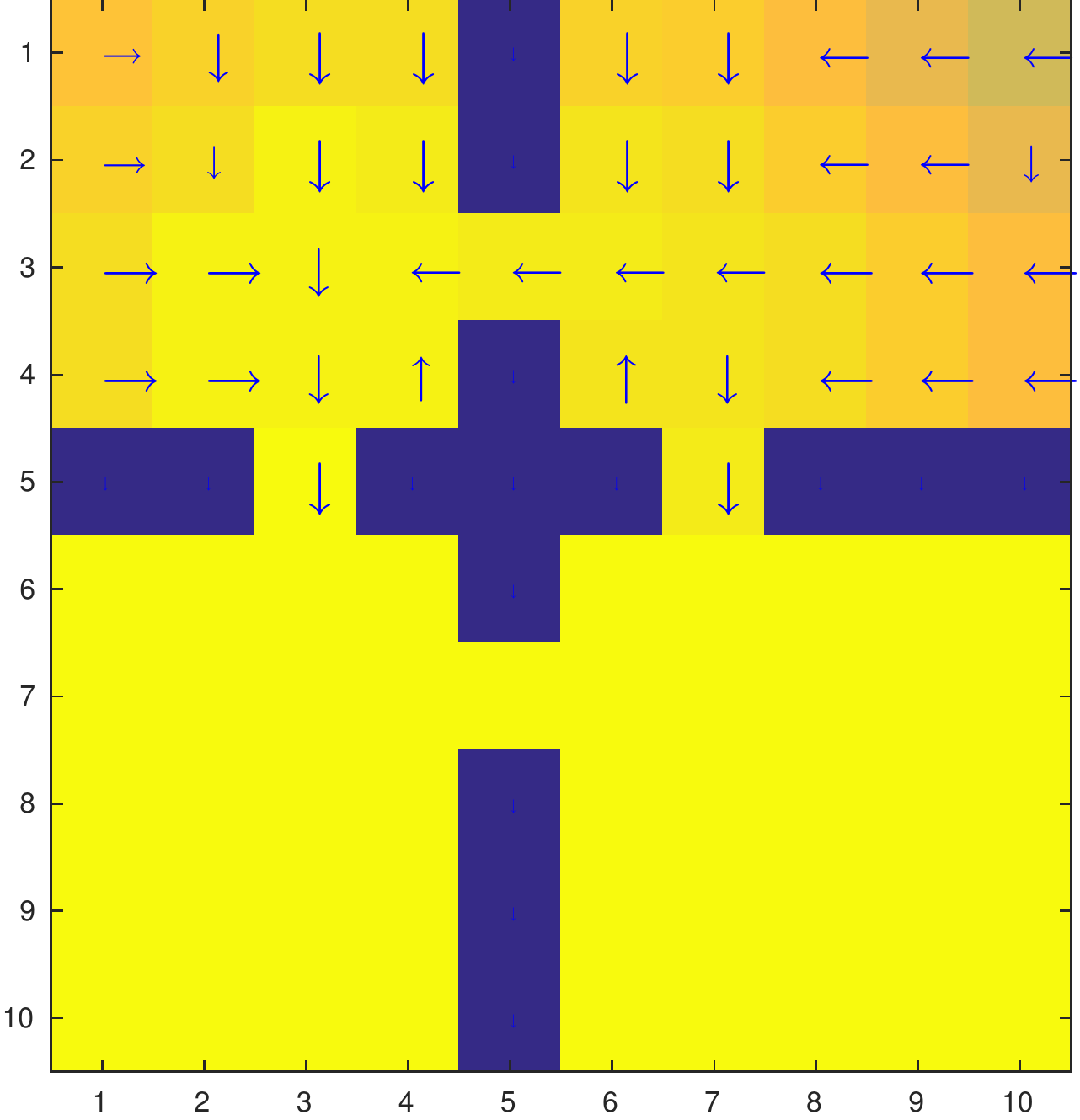}
\end{subfigure}
\hfill
\begin{subfigure}
	\centering
	\includegraphics[width = 0.2\textwidth]{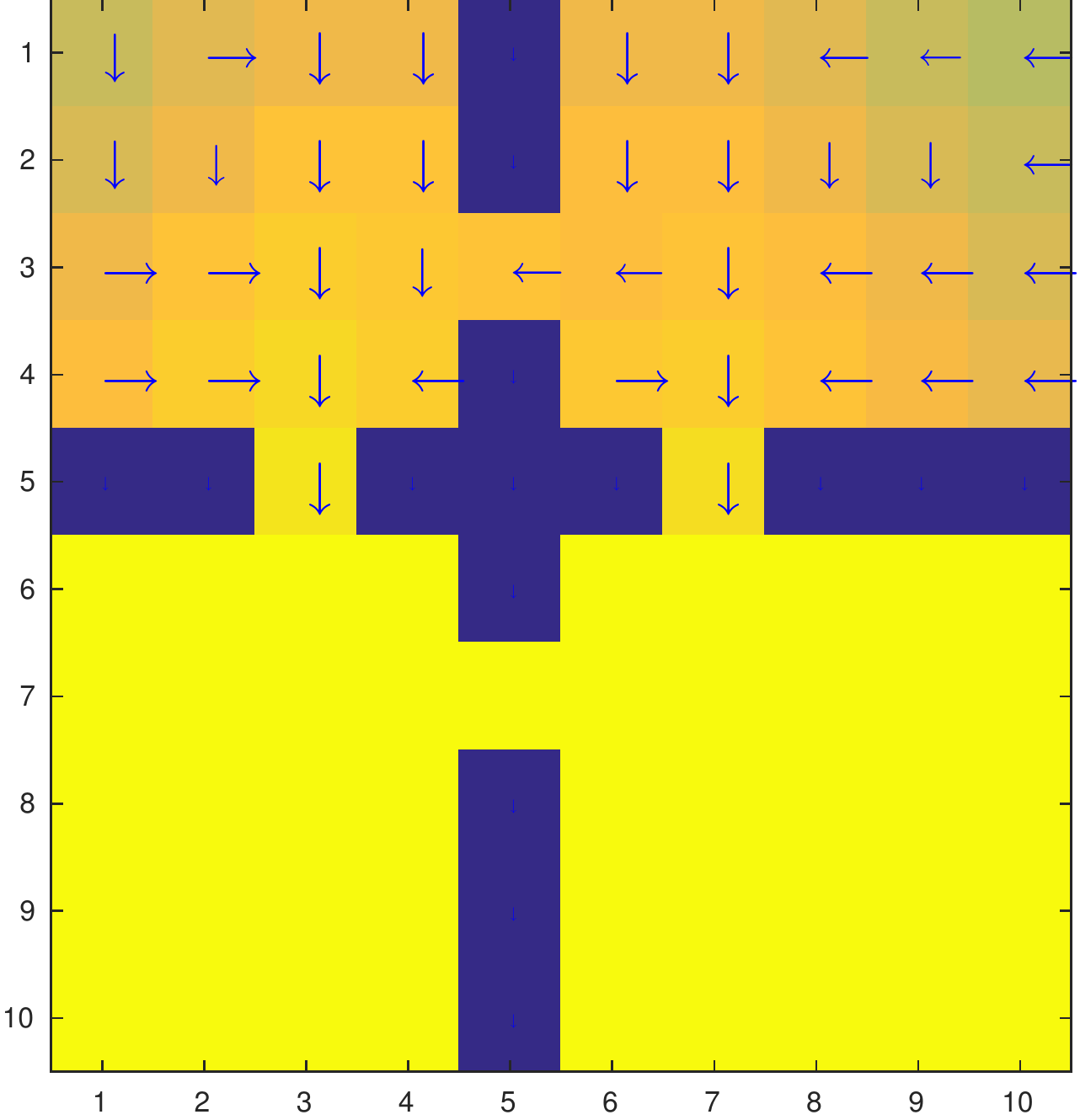}
\end{subfigure}
\hfill
\begin{subfigure}
	\centering
	\includegraphics[width = 0.2\textwidth]{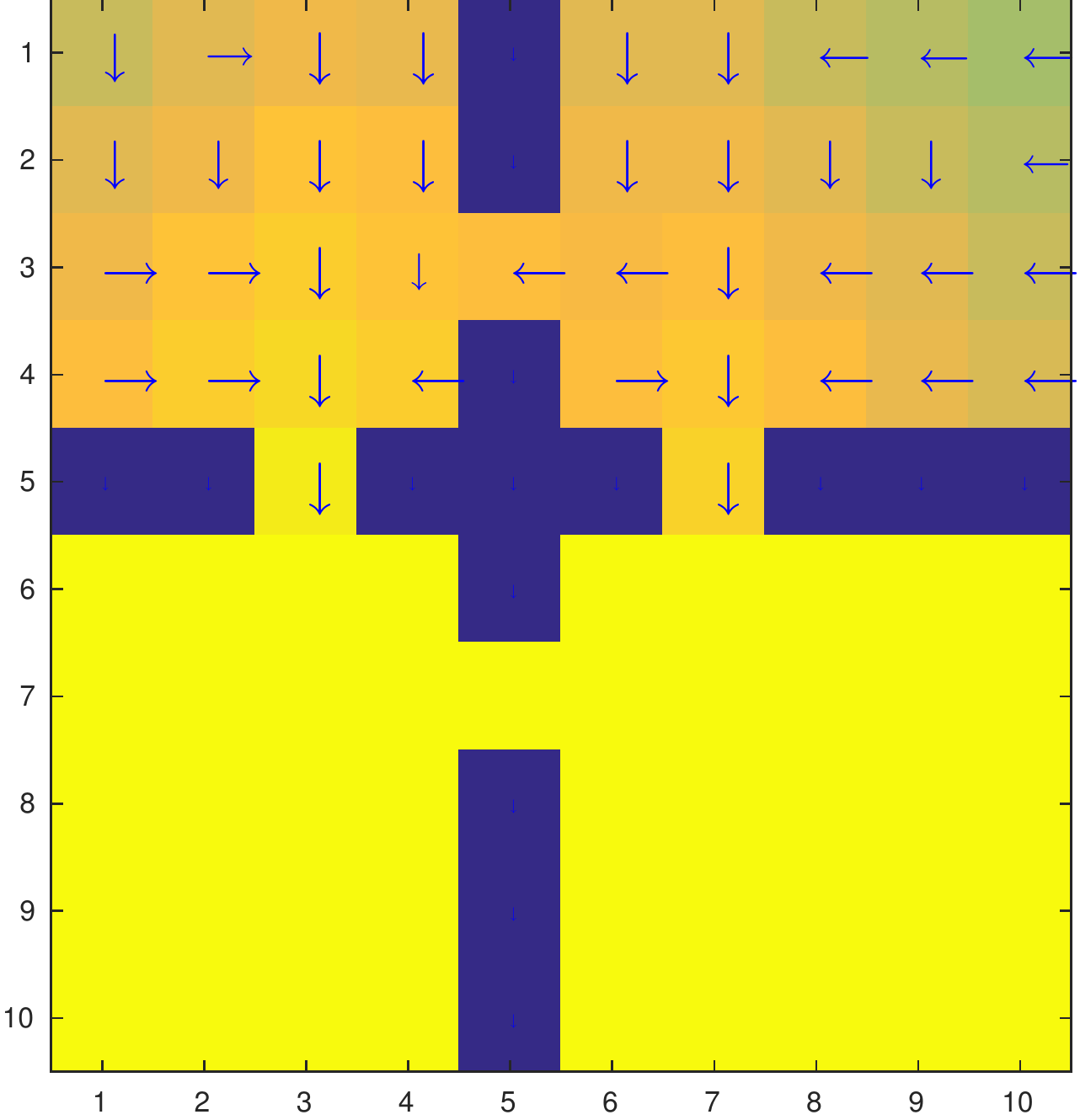}
\end{subfigure}
\hfill
\begin{subfigure}
	\centering
	\includegraphics[width = 0.2\textwidth]{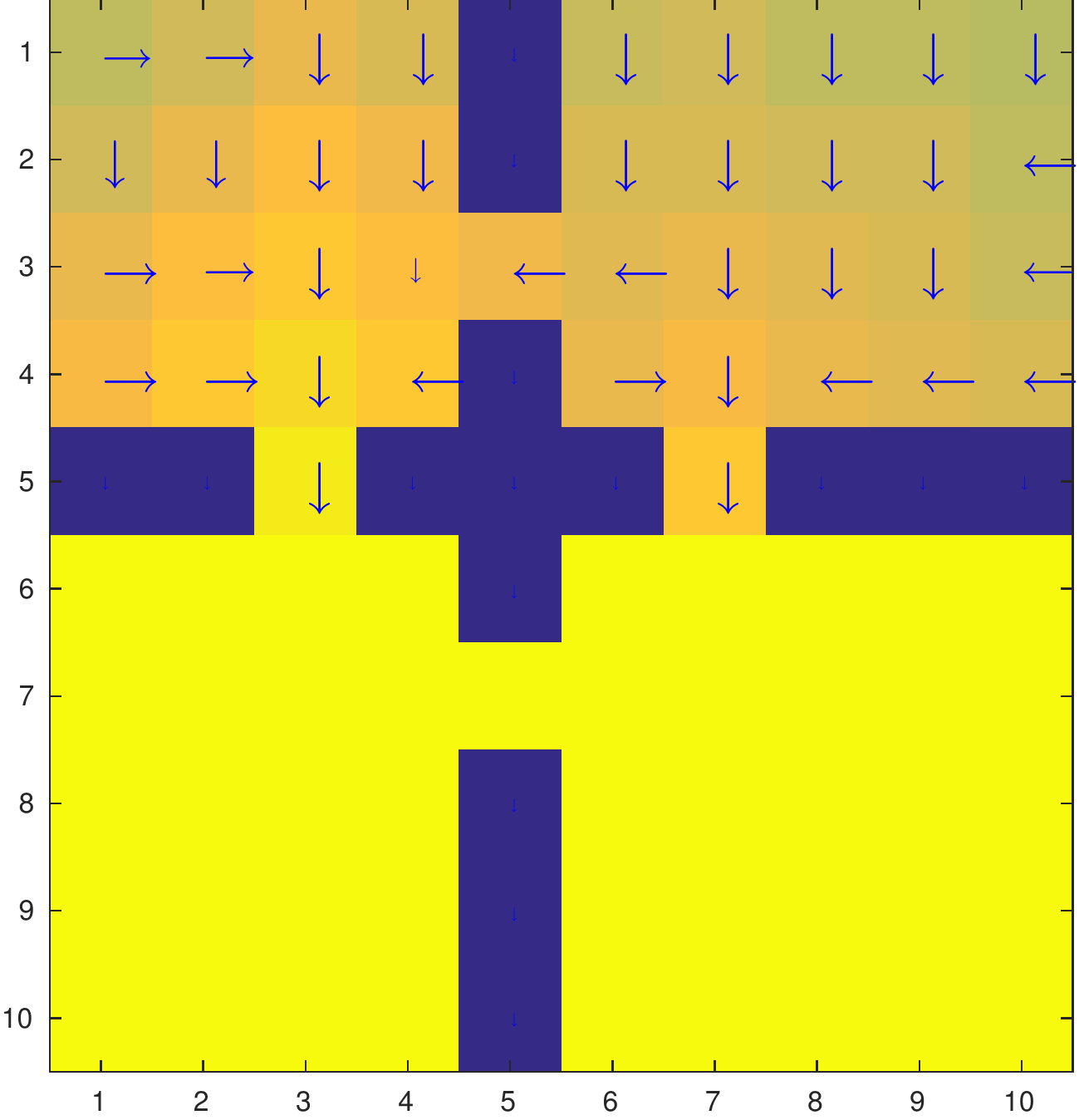}
\end{subfigure}
\begin{subfigure}
	\centering
	\includegraphics[width = 0.2\textwidth]{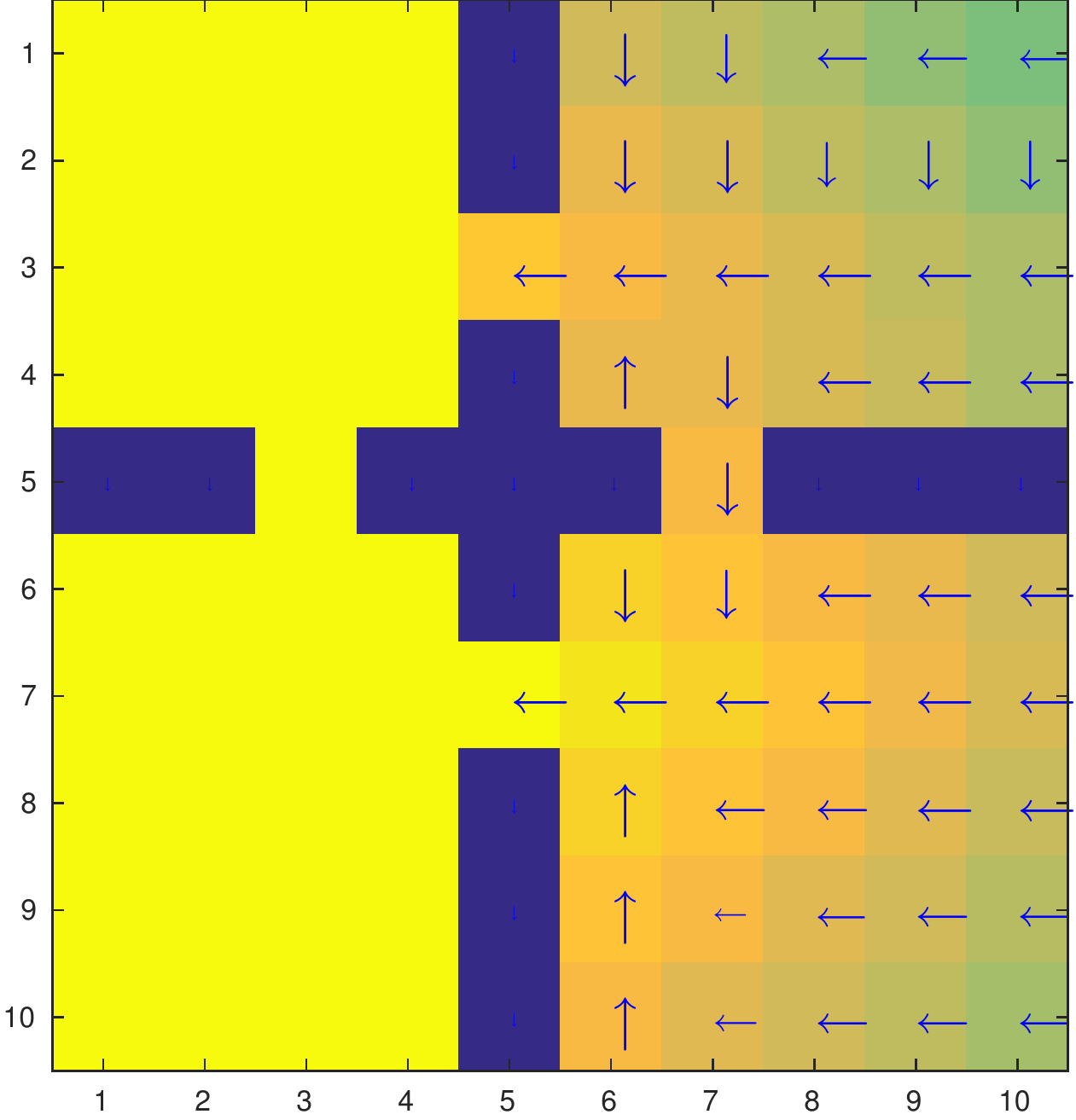}
\end{subfigure}
\hfill
\begin{subfigure}
	\centering
	\includegraphics[width = 0.2\textwidth]{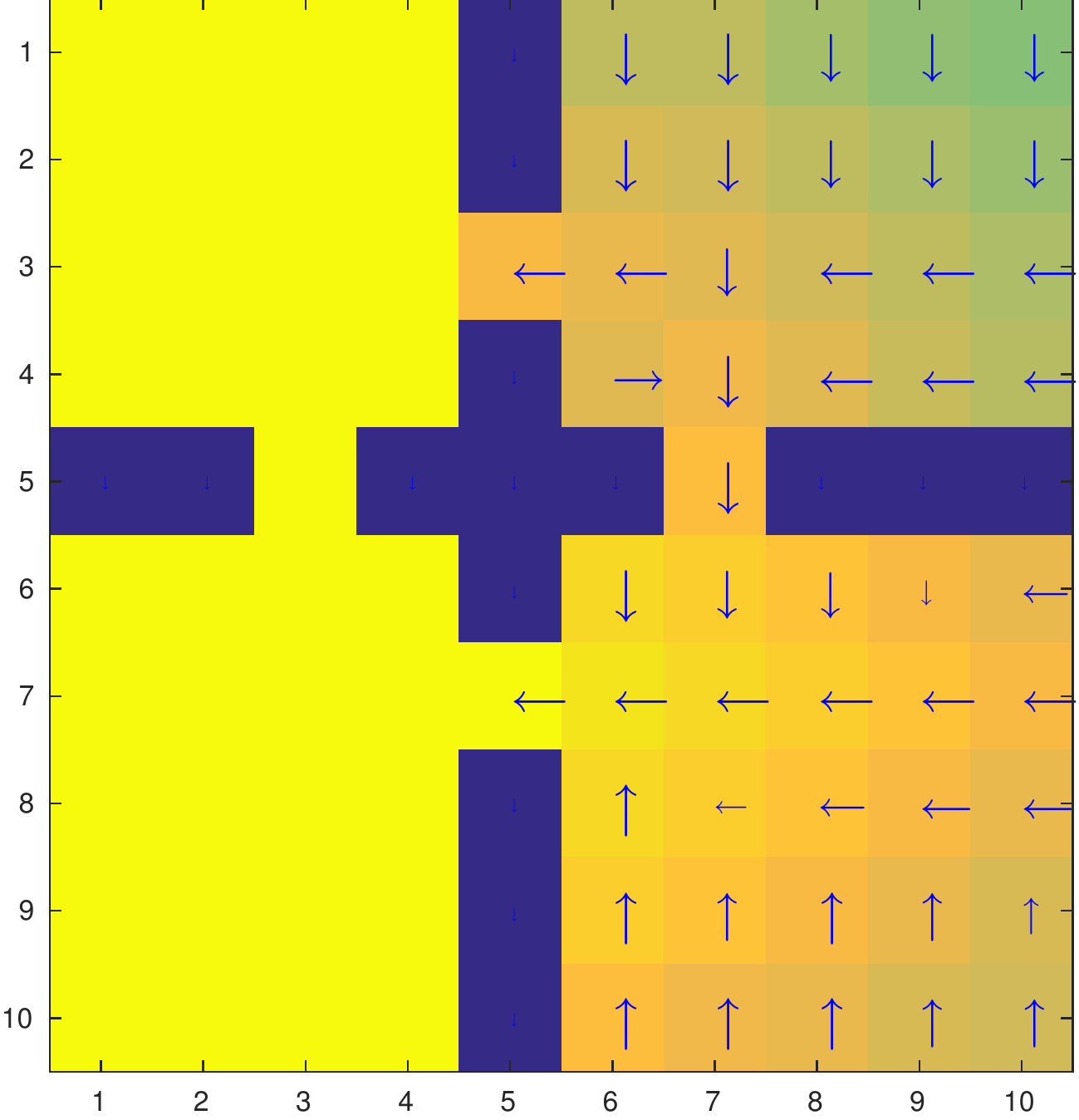}
\end{subfigure}
\hfill
\begin{subfigure}
	\centering
	\includegraphics[width = 0.2\textwidth]{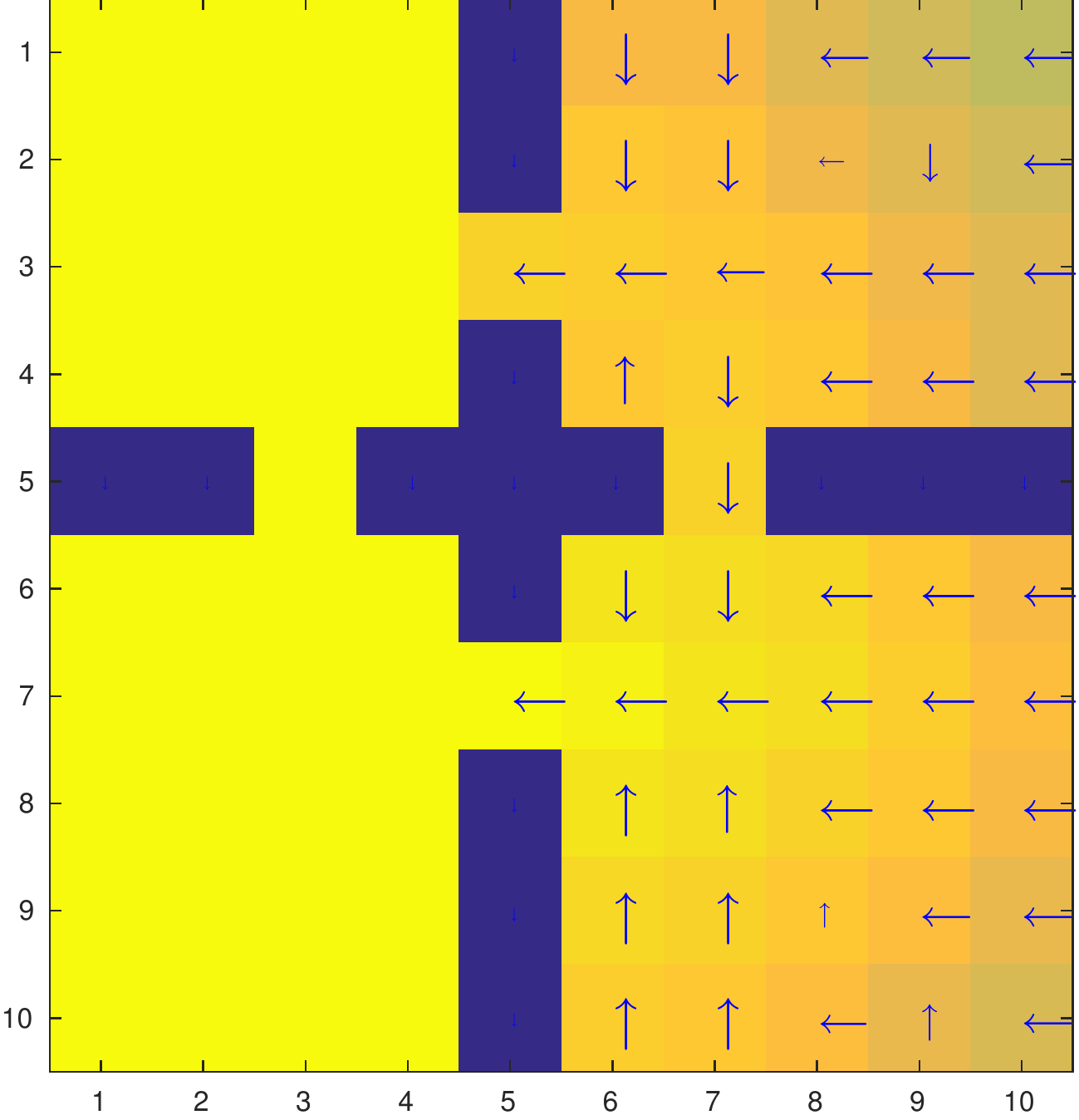}
\end{subfigure}
\hfill
\begin{subfigure}
	\centering
	\includegraphics[width = 0.2\textwidth]{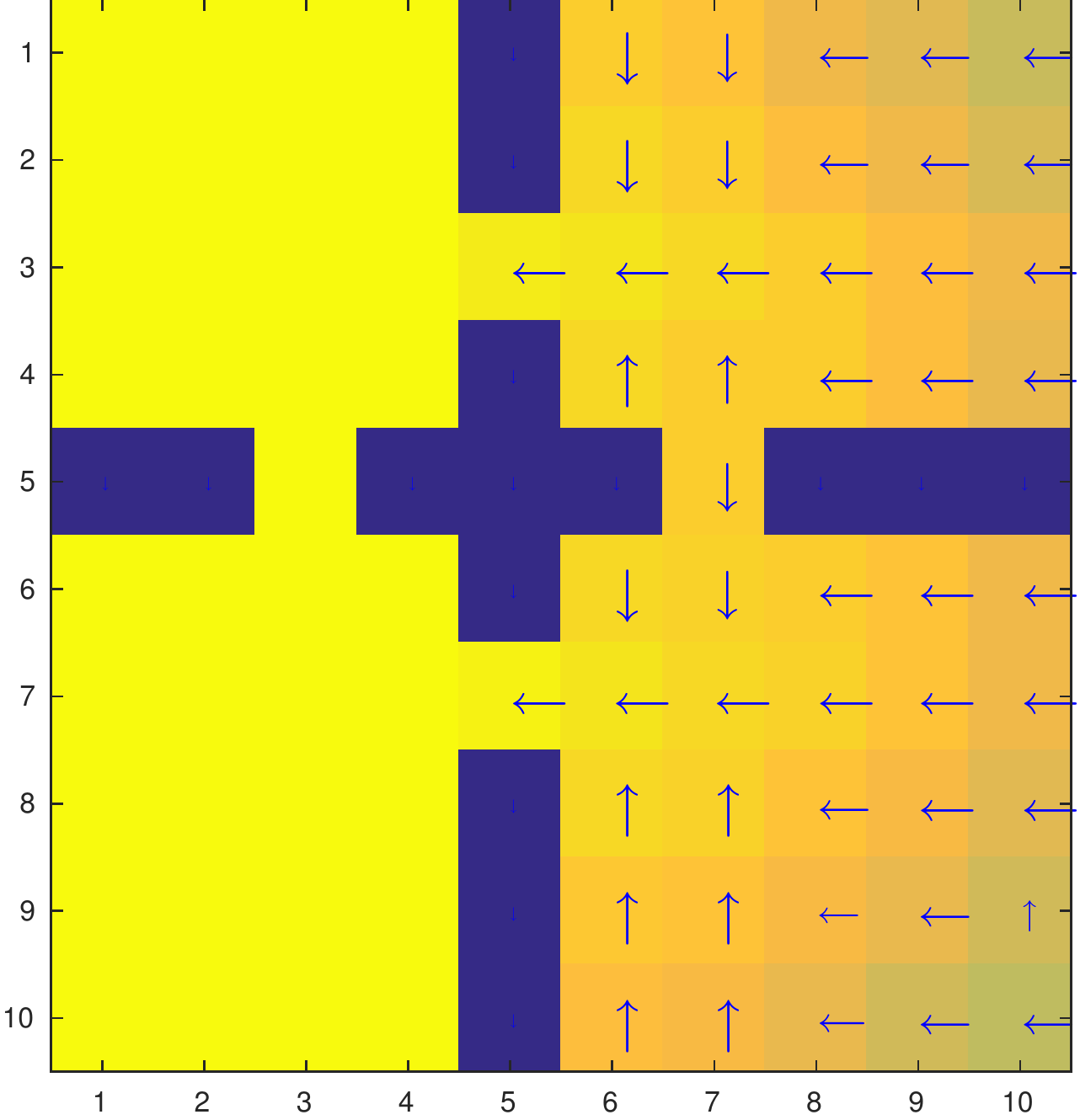}
\end{subfigure}
\begin{subfigure}
	\centering
	\includegraphics[width = 0.2\textwidth, height = 0.2\textwidth]{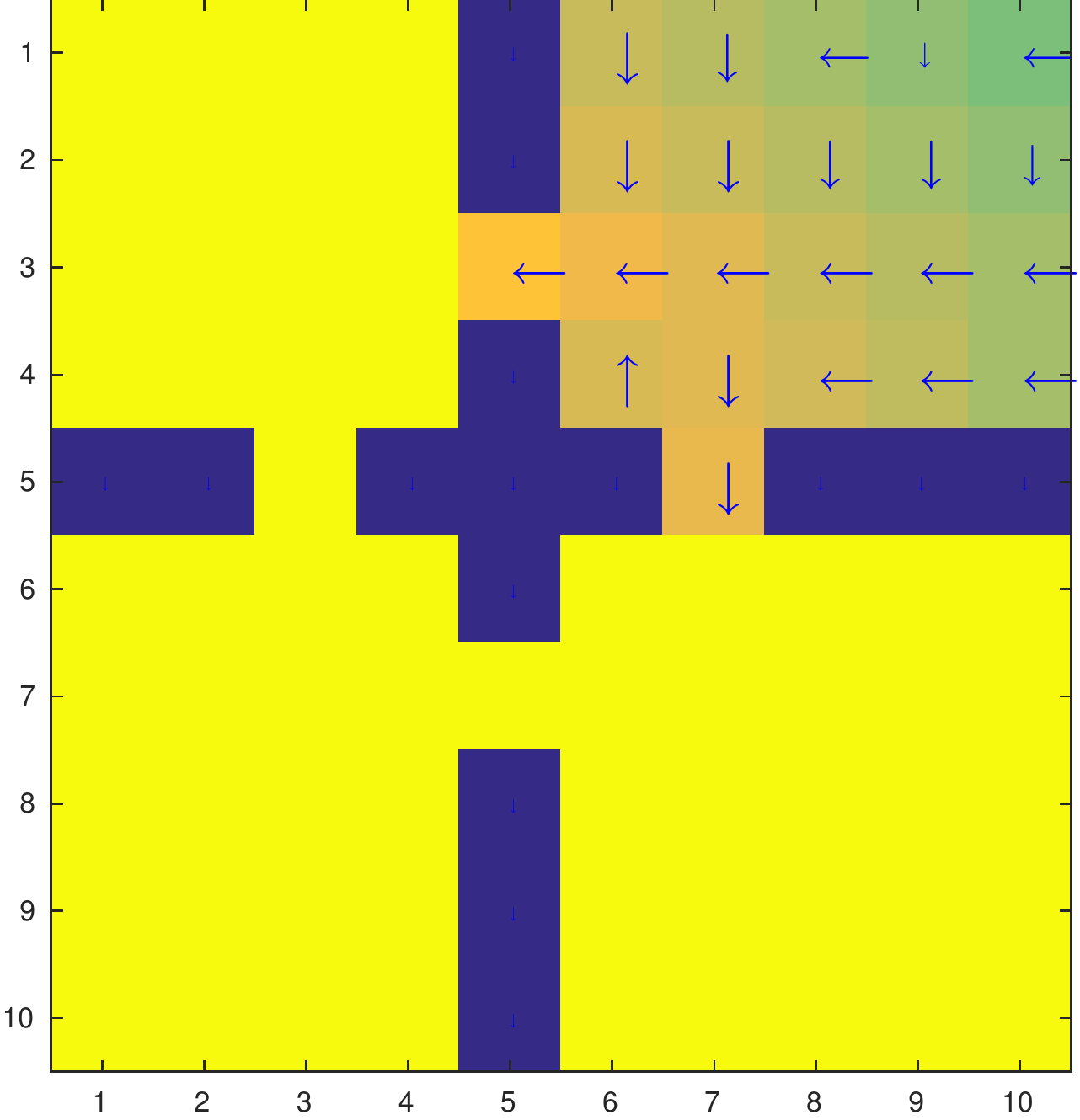}
\end{subfigure}
\hfill
\begin{subfigure}
	\centering
	\includegraphics[width = 0.2\textwidth, height = 0.2\textwidth]{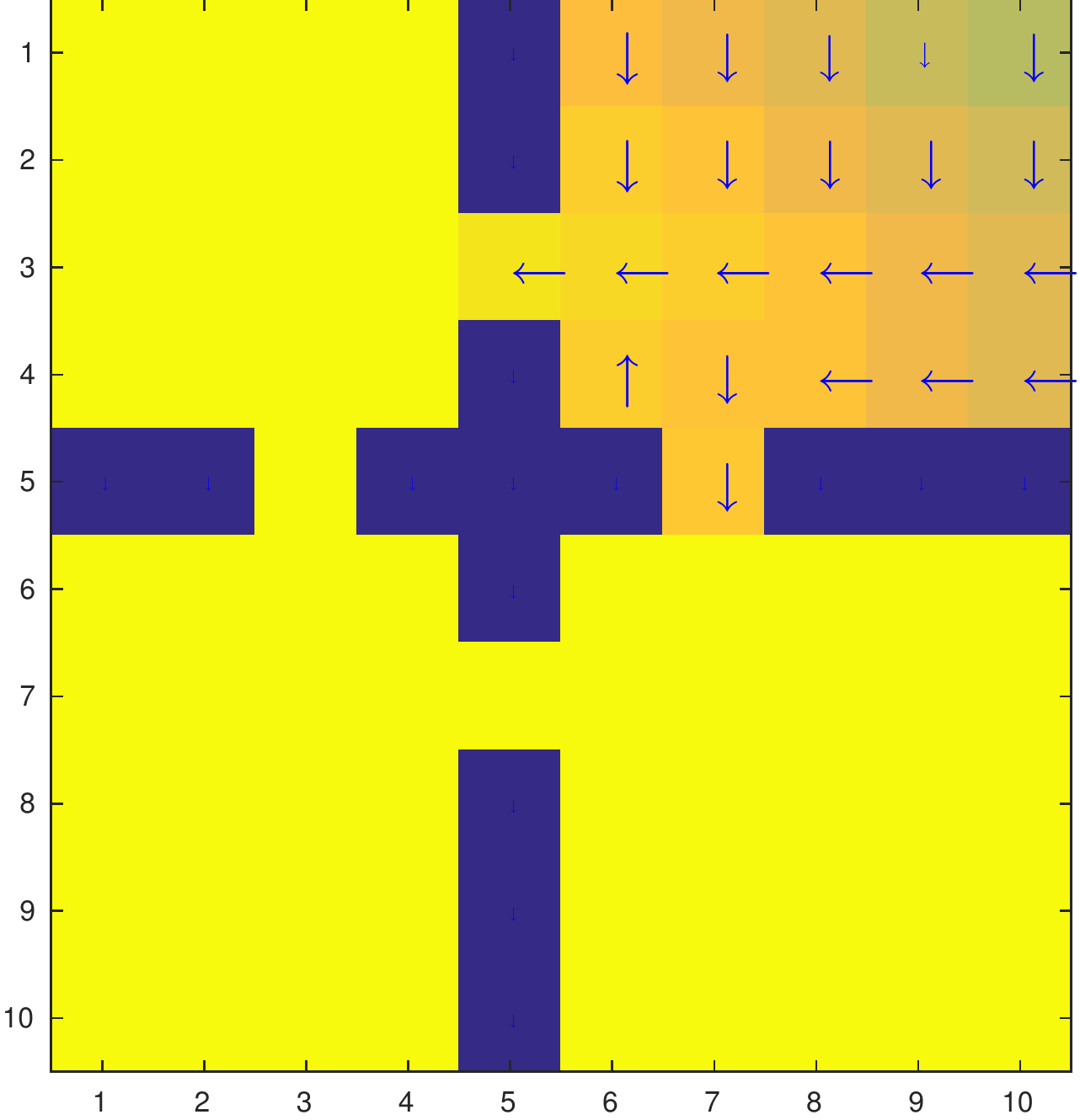}
\end{subfigure}
\hfill
\begin{subfigure}
	\centering
	\includegraphics[width = 0.2\textwidth, height = 0.2\textwidth]{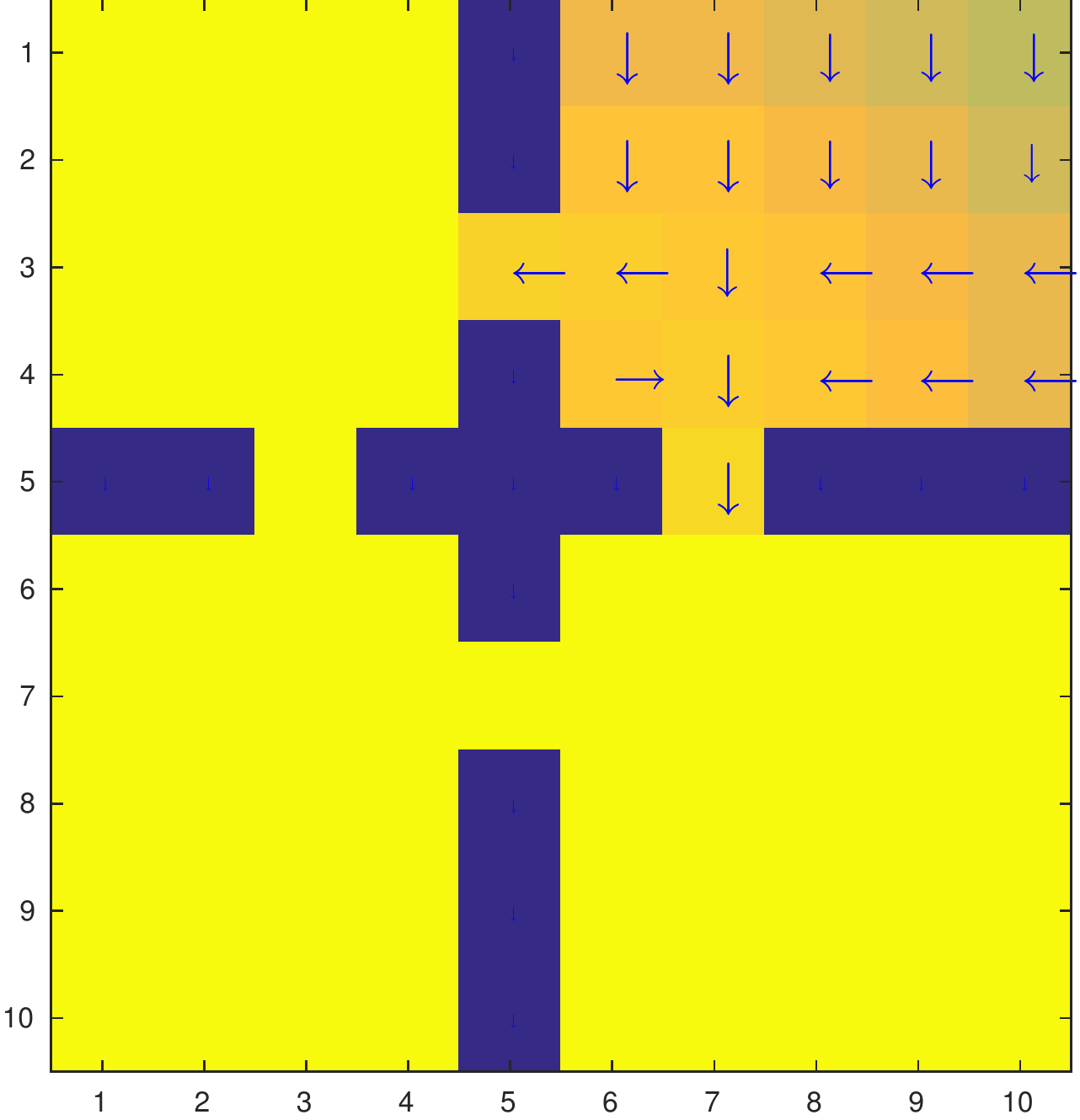}
\end{subfigure}
\hfill
\begin{subfigure}
	\centering
	\includegraphics[width = 0.2\textwidth, height = 0.2\textwidth]{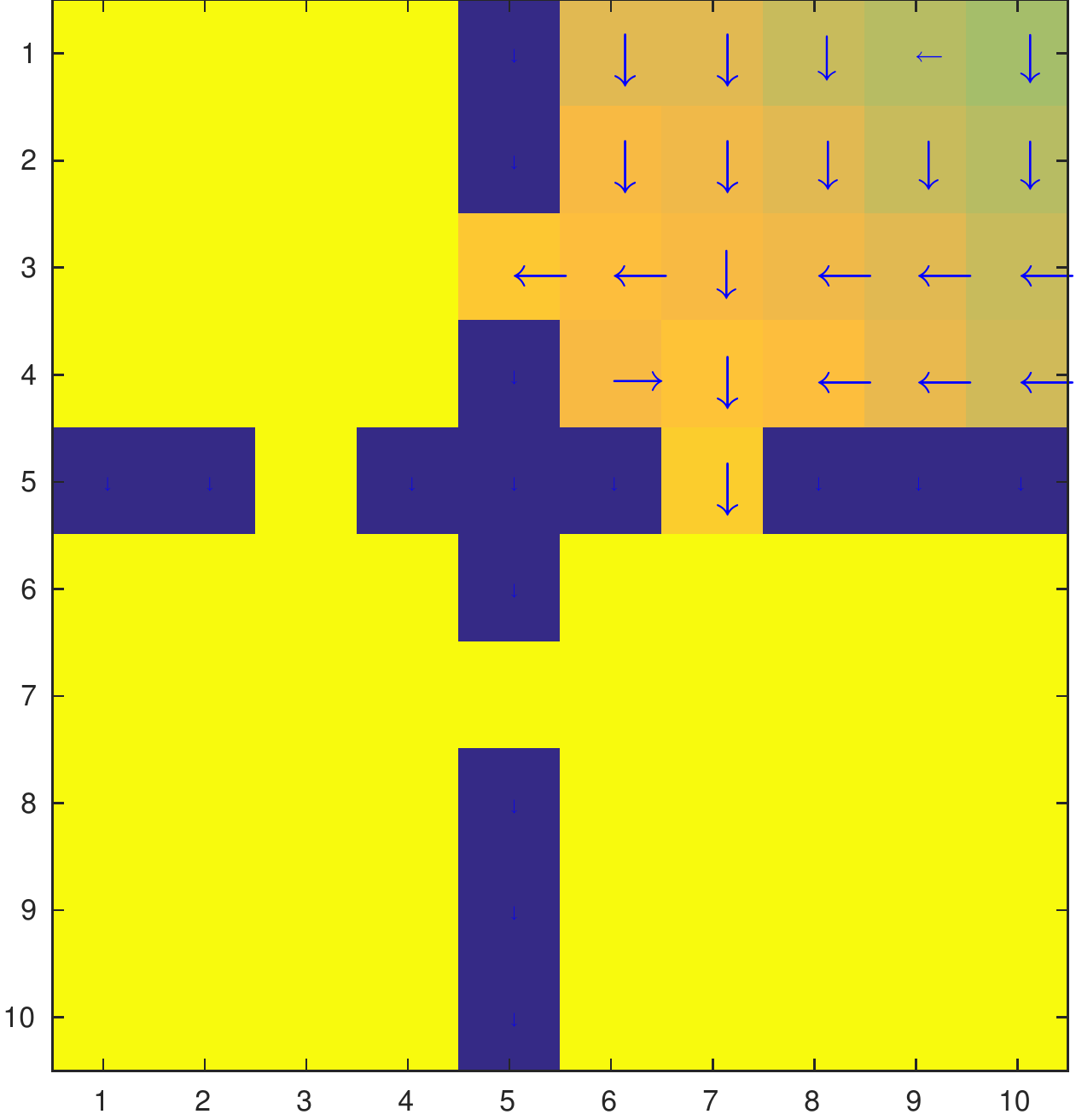}
\end{subfigure}
\begin{subfigure}
	\centering
	\includegraphics[width = 0.2\textwidth, height = 0.2\textwidth]{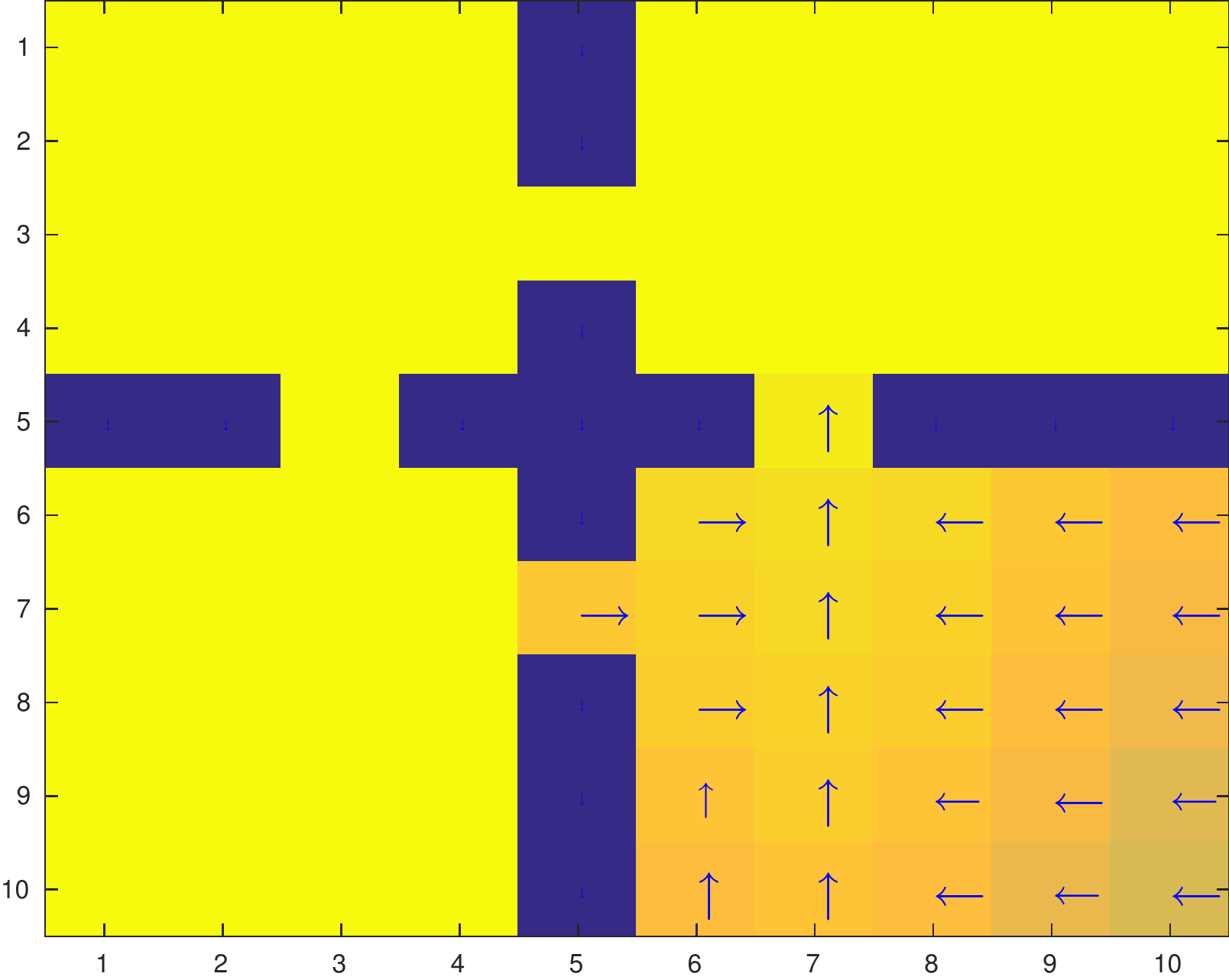}
\end{subfigure}
\hfill
\begin{subfigure}
	\centering
	\includegraphics[width = 0.2\textwidth, height = 0.2\textwidth]{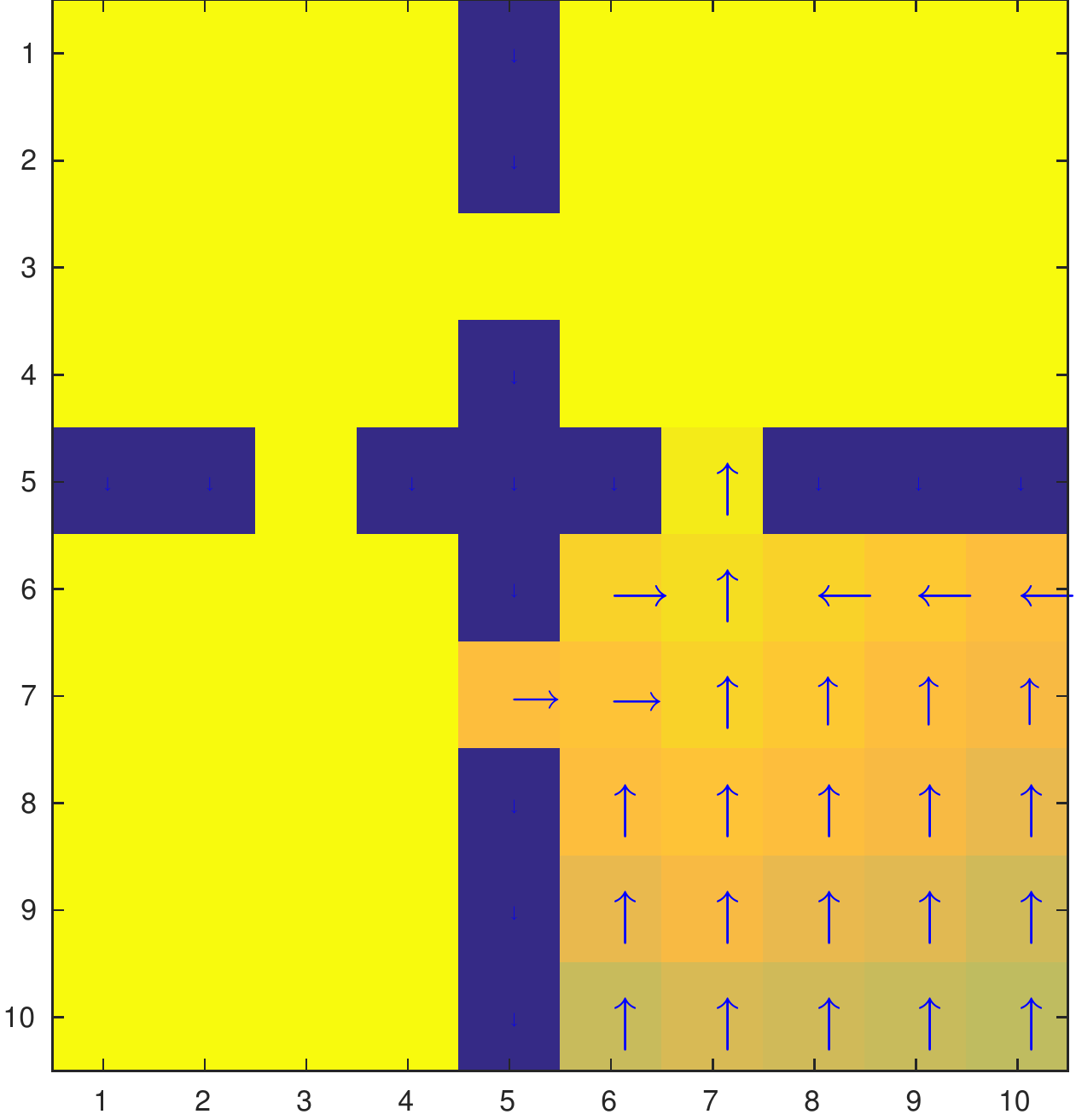}
\end{subfigure}
\hfill
\begin{subfigure}
	\centering
	\includegraphics[width = 0.2\textwidth, height = 0.2\textwidth]{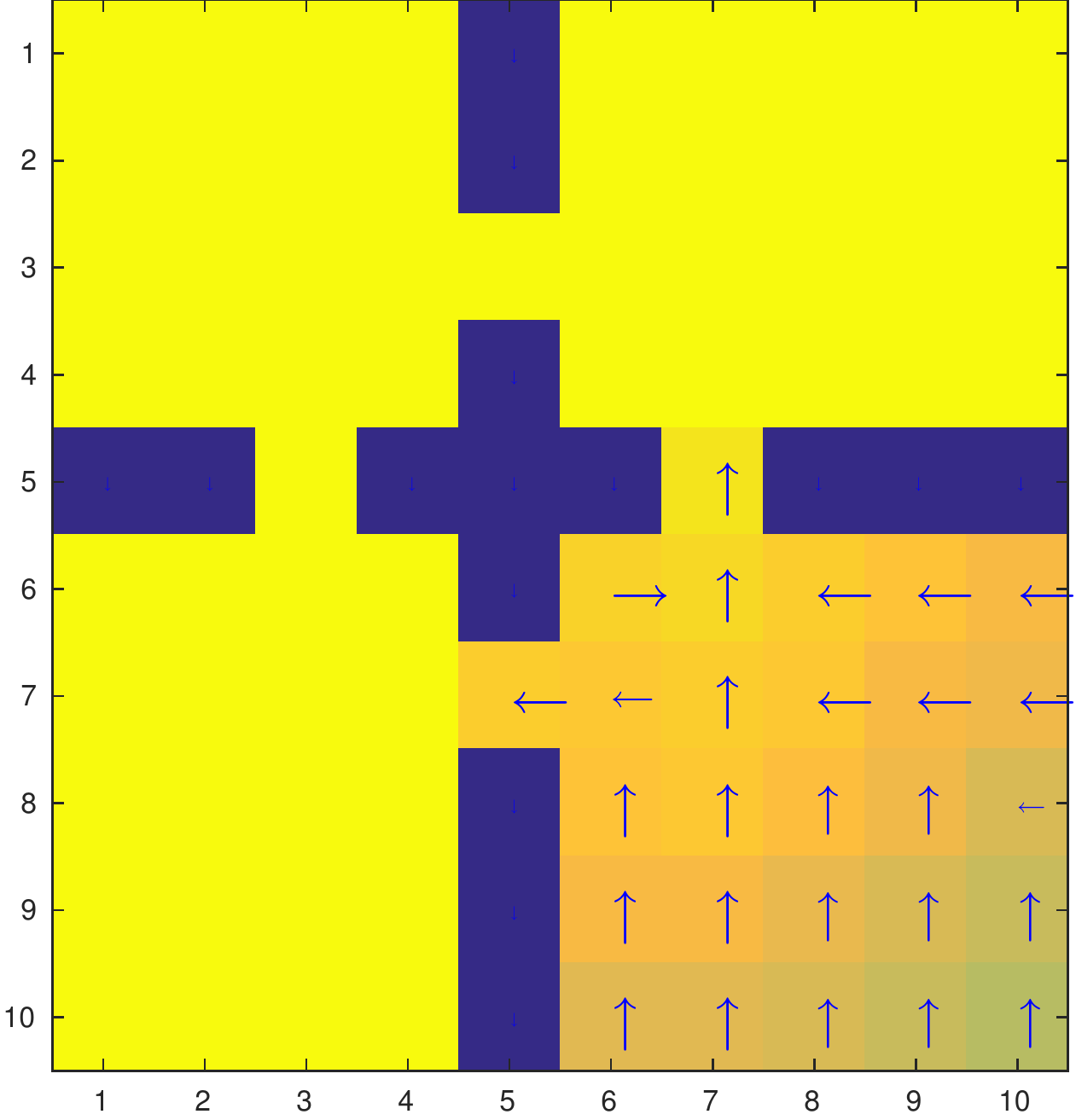}
\end{subfigure}
\hfill
\begin{subfigure}
	\centering
	\includegraphics[width = 0.2\textwidth, height = 0.2\textwidth]{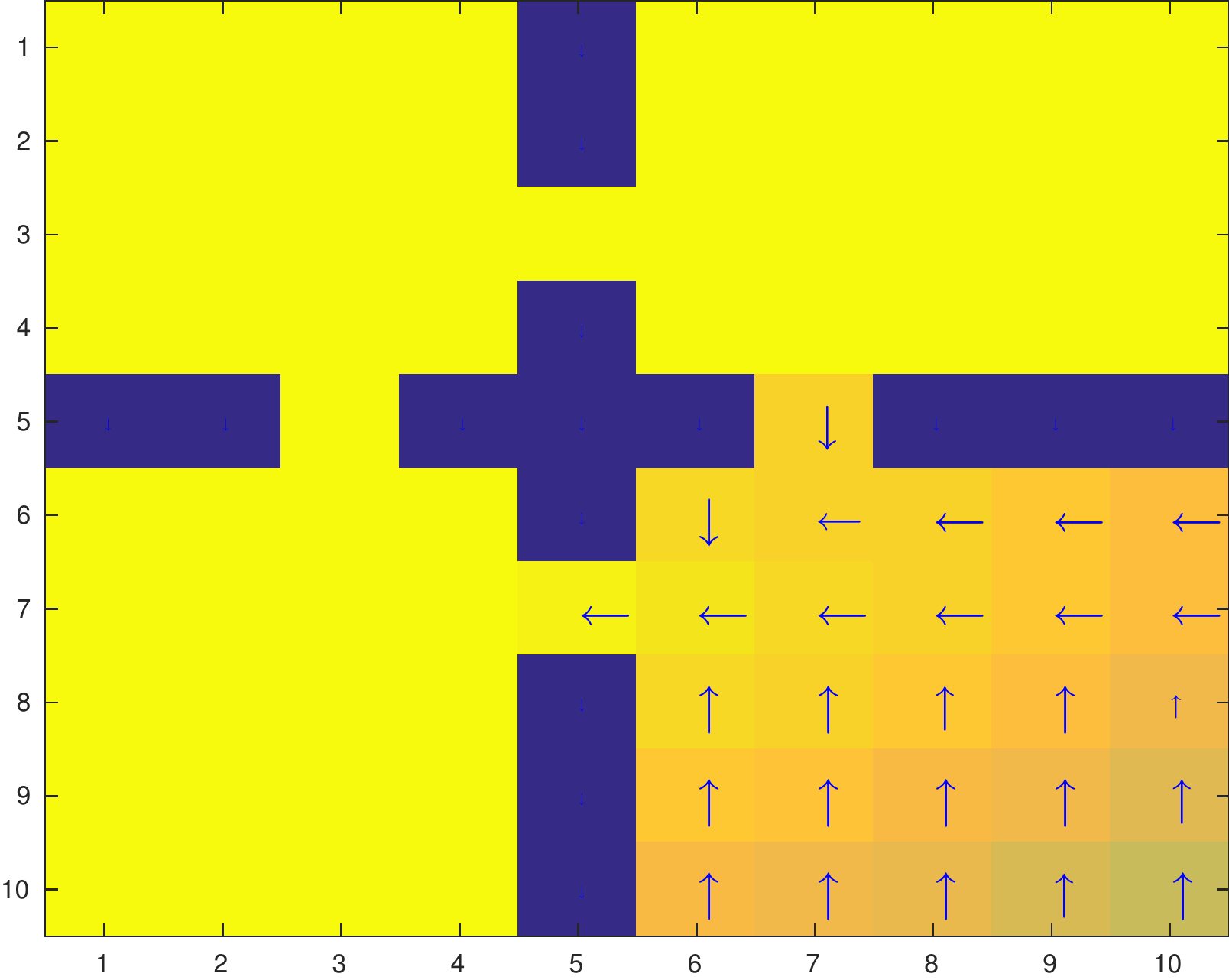}
\end{subfigure}
\begin{subfigure}
	\centering
	\includegraphics[width = 0.2\textwidth, height = 0.2\textwidth]{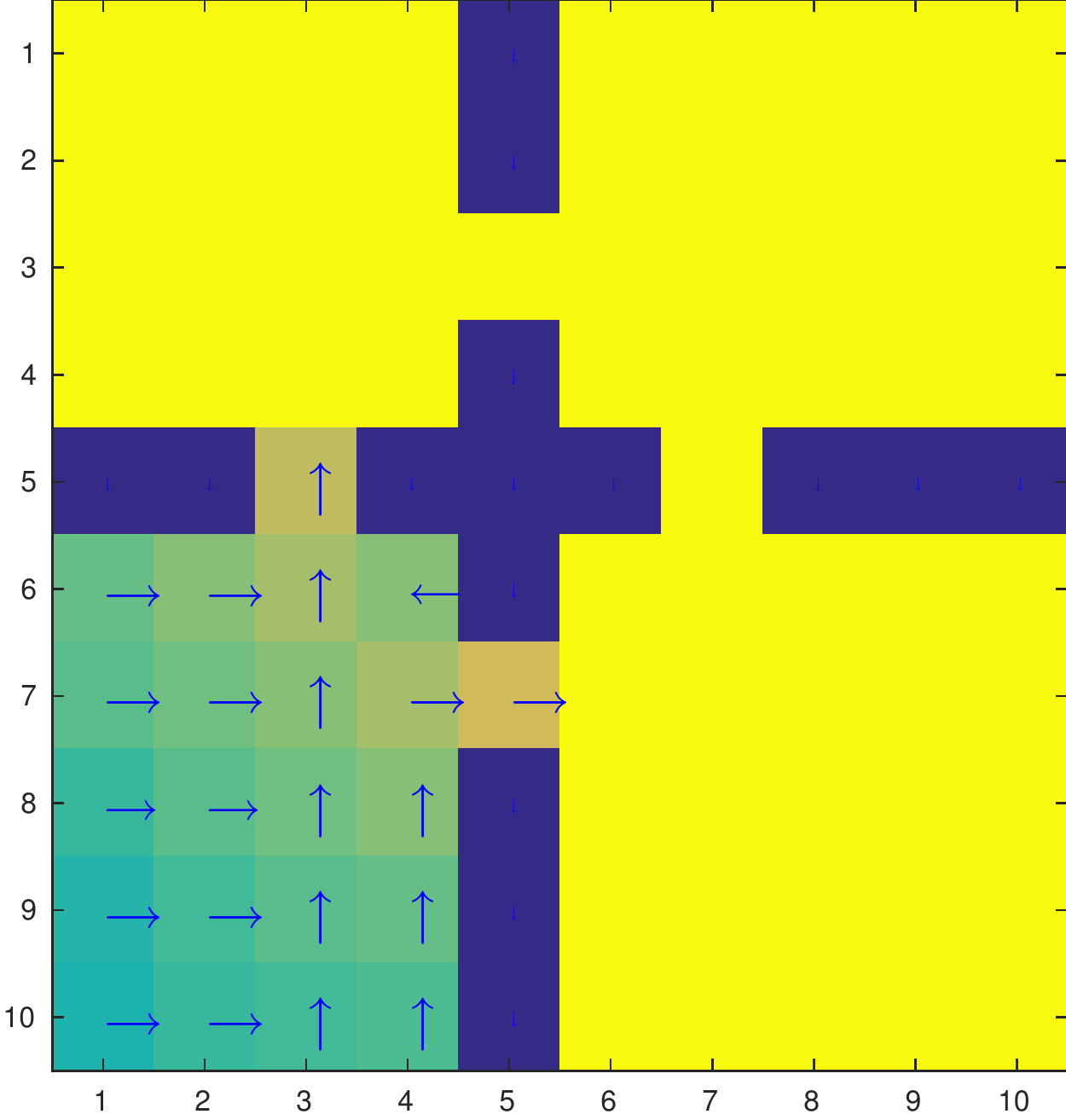}
\end{subfigure}
\hfill
\begin{subfigure}
	\centering
	\includegraphics[width = 0.2\textwidth, height = 0.2\textwidth]{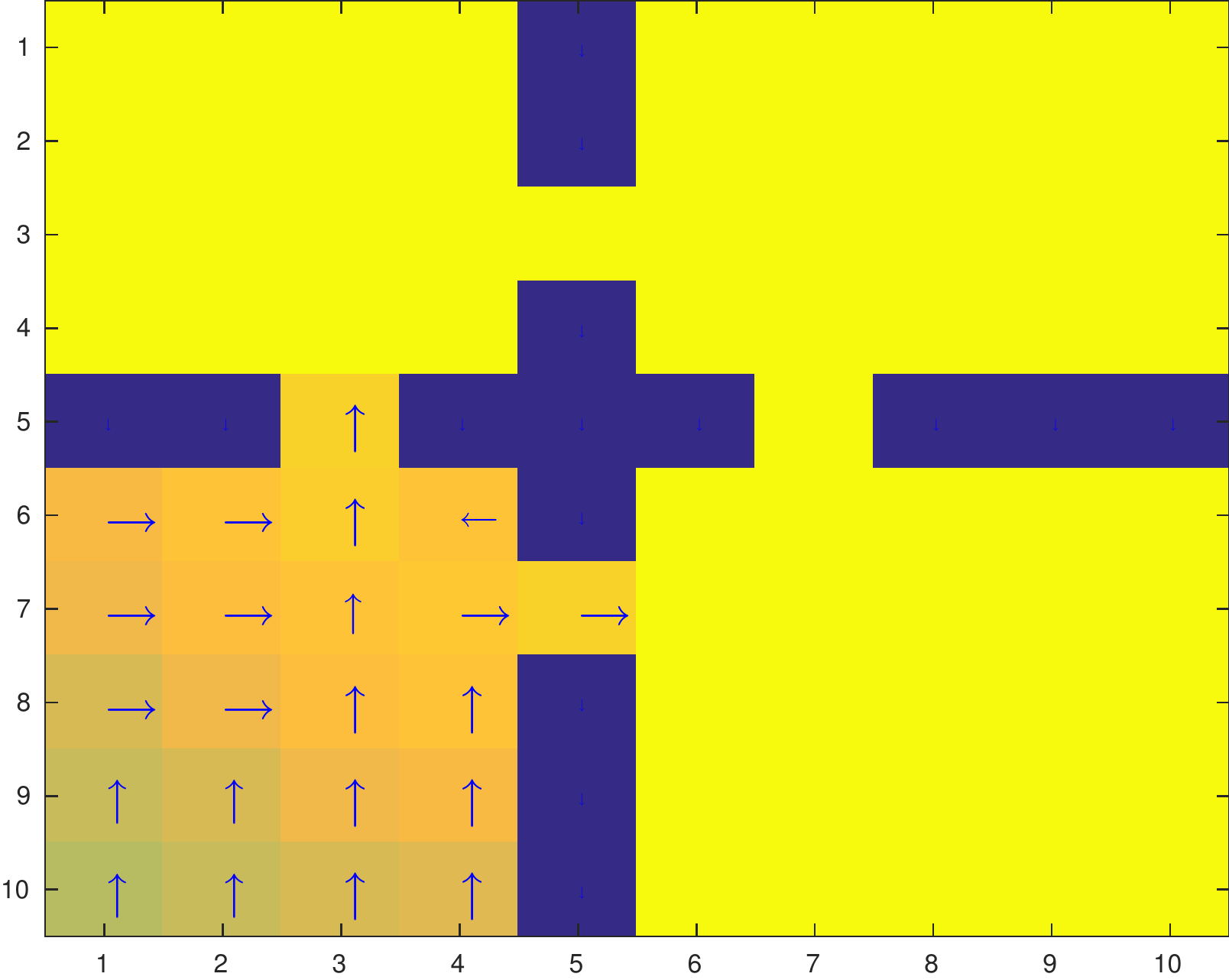}
\end{subfigure}
\hfill
\begin{subfigure}
	\centering
	\includegraphics[width = 0.2\textwidth, height = 0.2\textwidth]{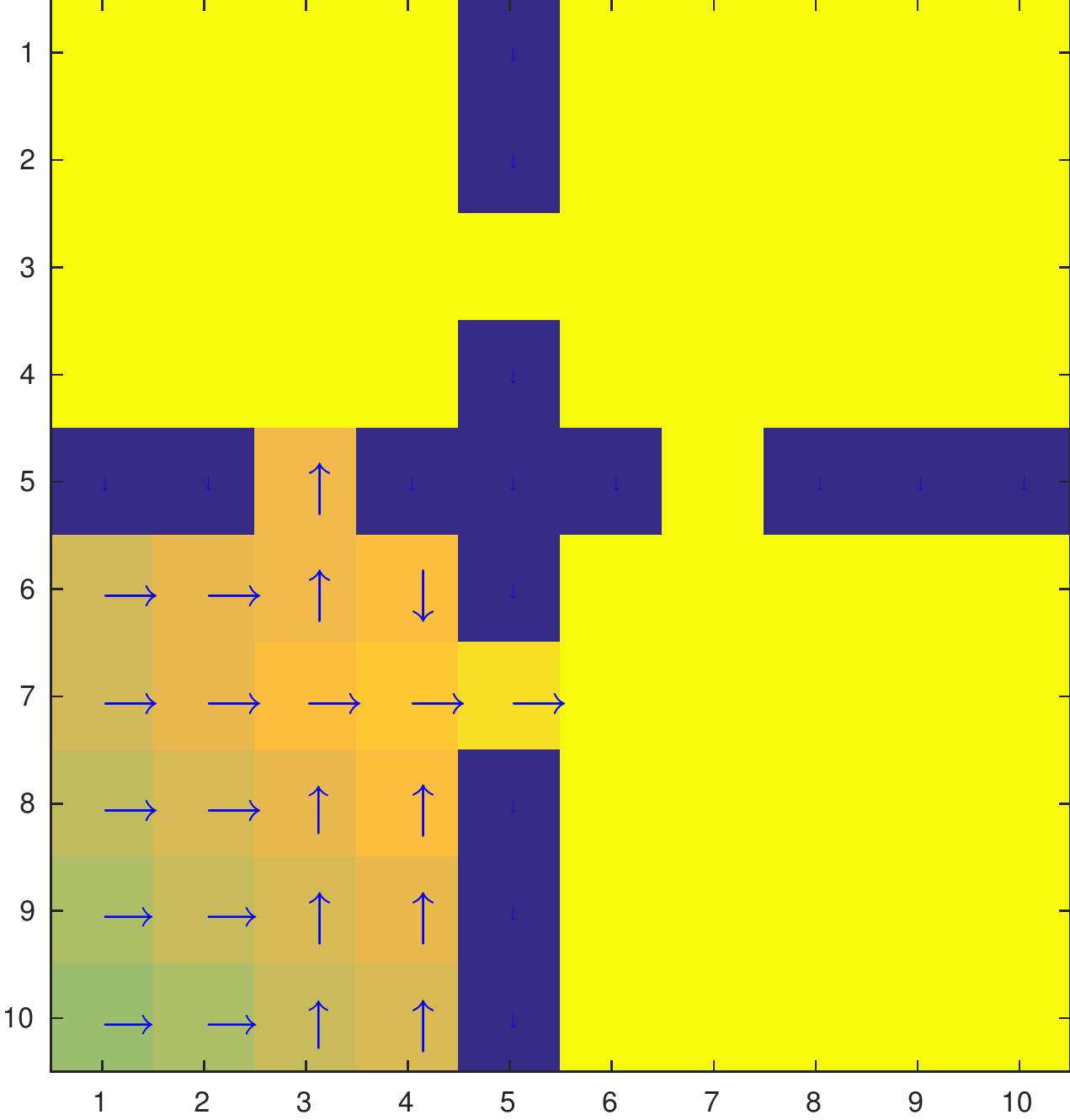}
\end{subfigure}
\hfill
\begin{subfigure}
	\centering
	\includegraphics[width = 0.2\textwidth, height = 0.2\textwidth]{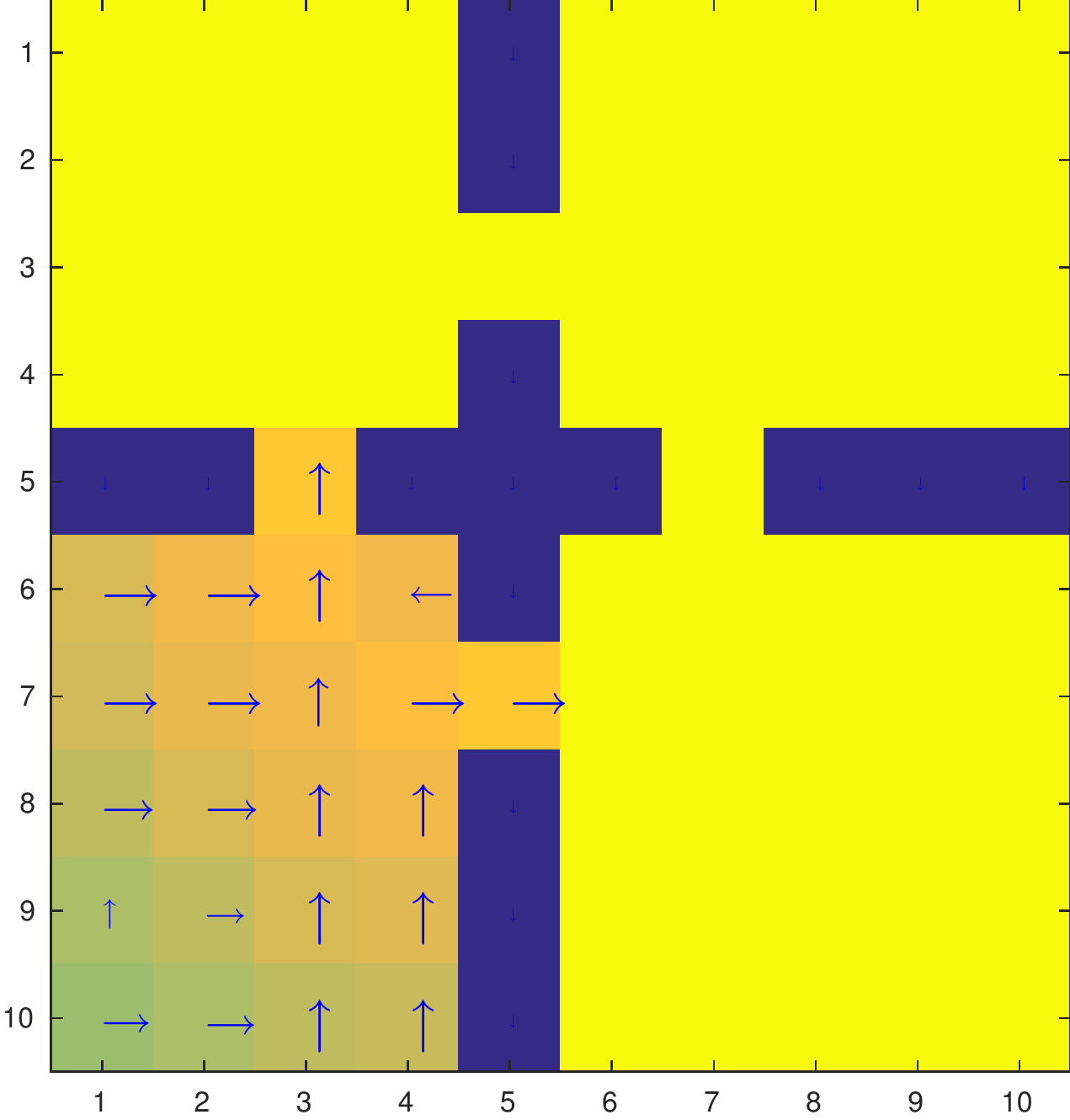}
\end{subfigure}
\begin{subfigure}
	\centering
	\includegraphics[width = 0.2\textwidth, height = 0.2\textwidth]{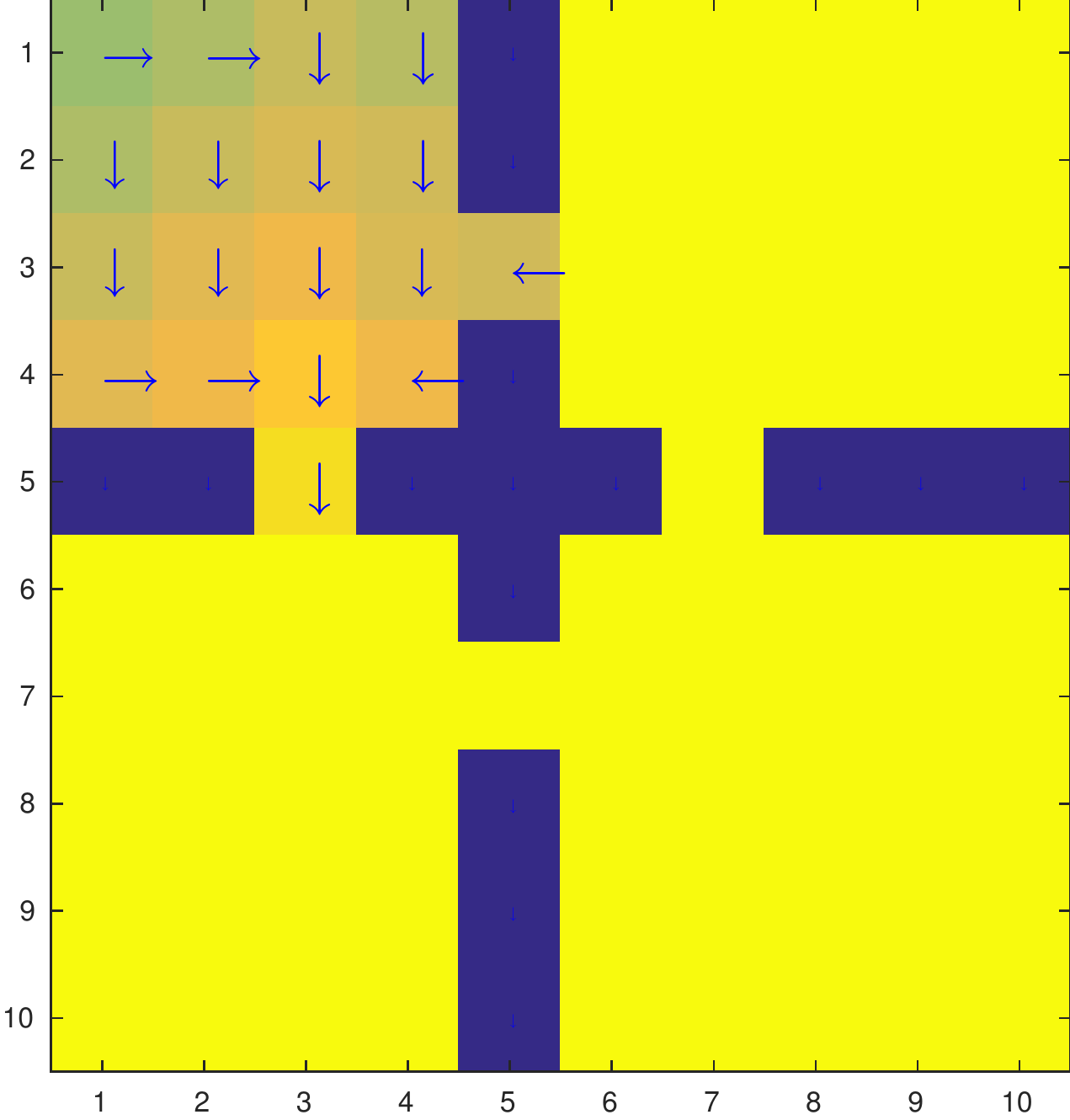}
\end{subfigure}
\hfill
\begin{subfigure}
	\centering
	\includegraphics[width = 0.2\textwidth, height = 0.2\textwidth]{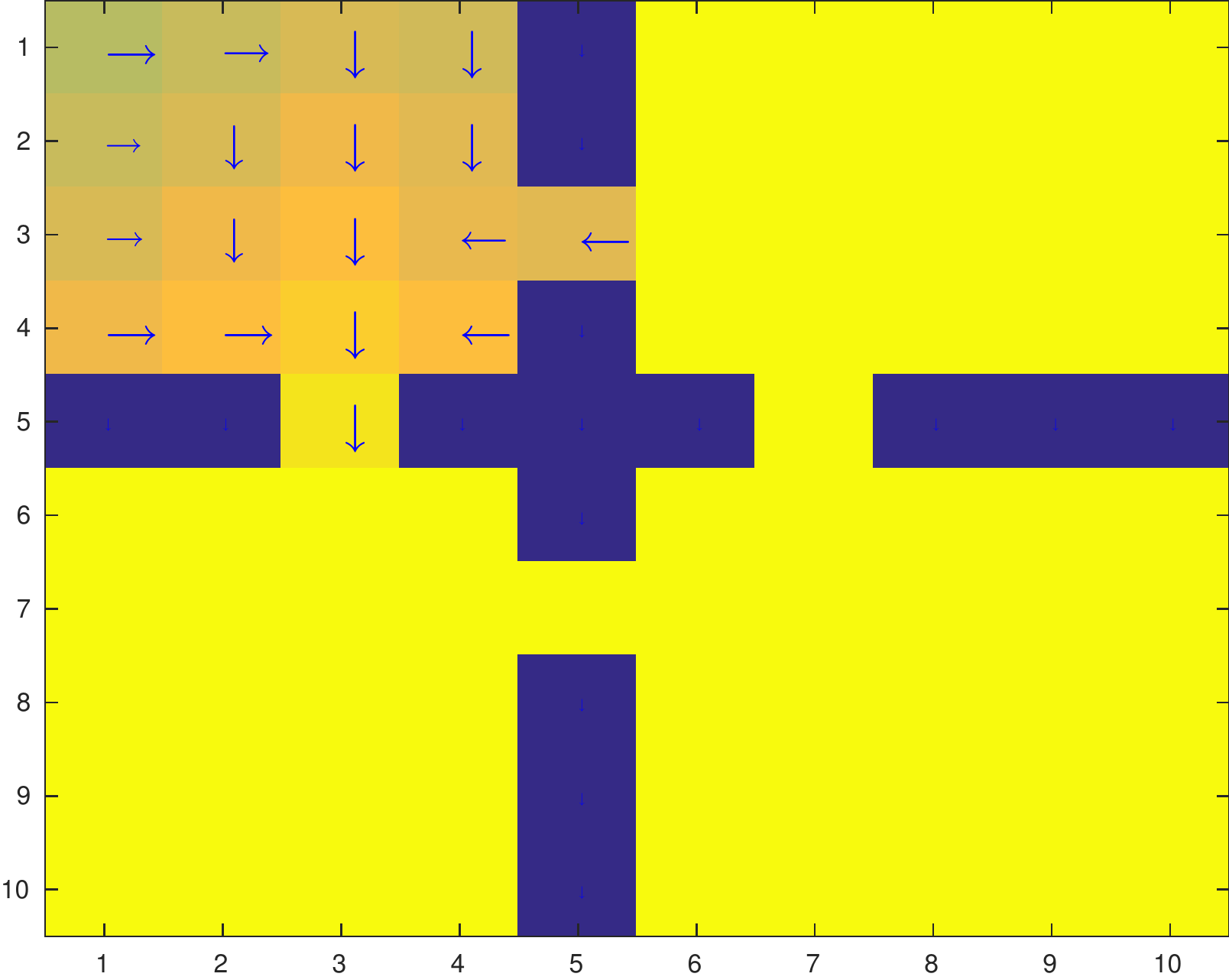}
\end{subfigure}
\hfill
\begin{subfigure}
	\centering
	\includegraphics[width = 0.2\textwidth, height = 0.2\textwidth]{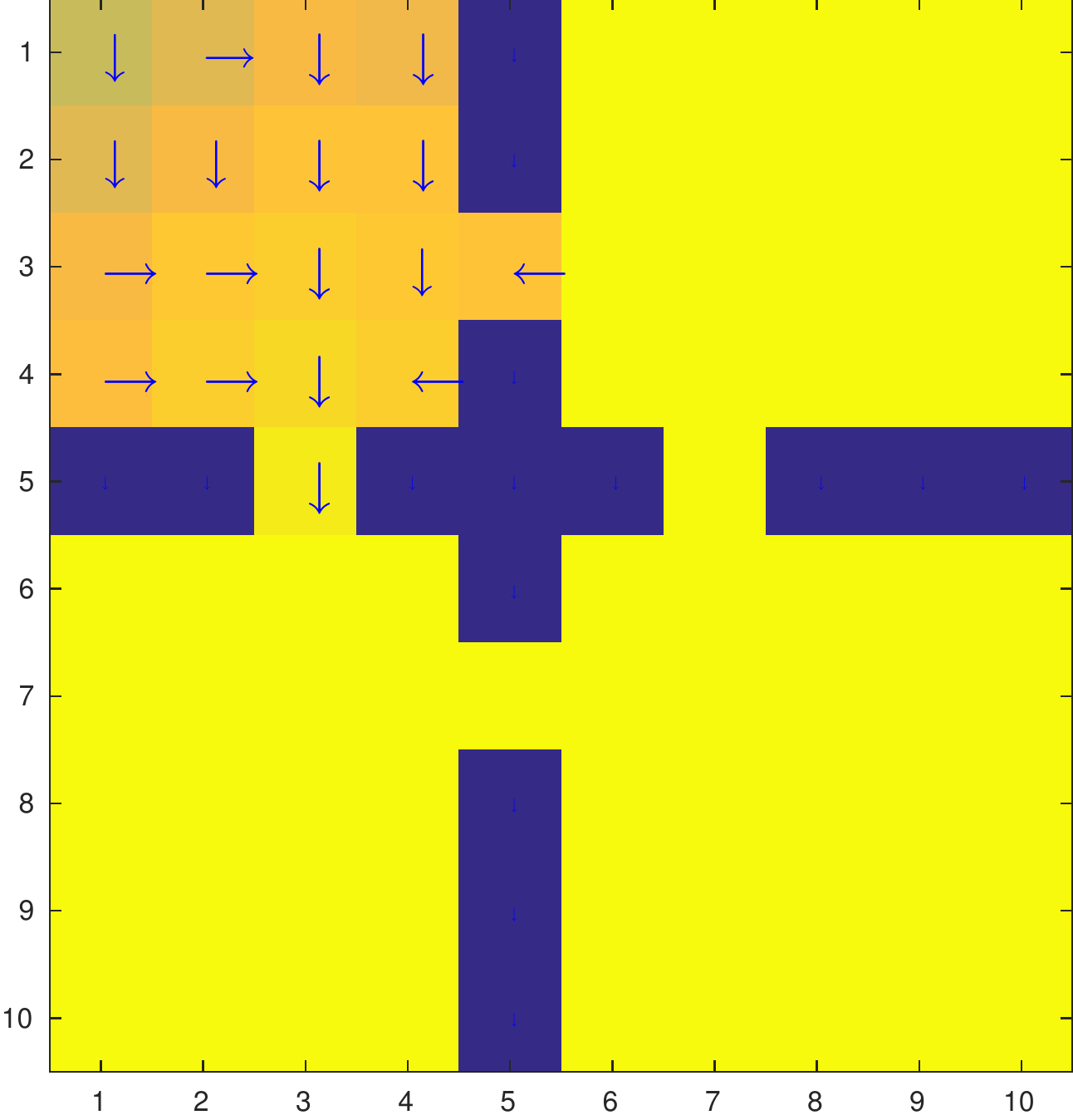}
\end{subfigure}
\hfill
\begin{subfigure}
	\centering
	\includegraphics[width = 0.2\textwidth, height = 0.2\textwidth]{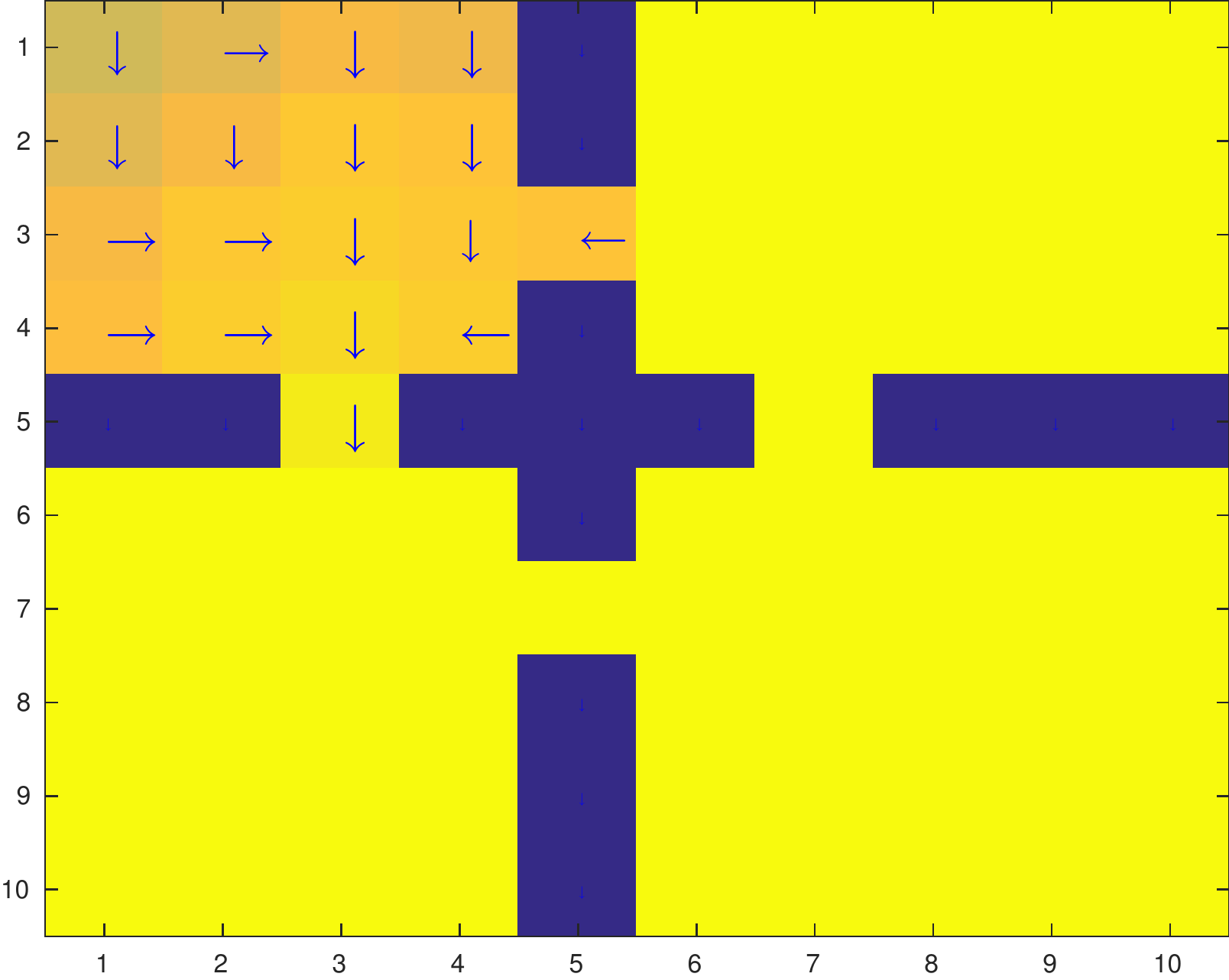}
\end{subfigure}
\caption{Learned greedy policies (as indicated by the arrows) and value functions (coloring indicates the value $V(s) = \max_a(Q(s,a))$ ) enabling navigation to any of the four rooms, based only on the share feature subspace discovered in the multi-task value function learning of $30$ goals randomly sampled in the environment. The value functions were learnt using (single-task) FQI on top of features $\psi^{ASO}_s$ and we show the results when using $30$, $50$, $100$ and respectively $300$ samples from the option-defined MDP.}
\label{fig:options_rooms_all}
\end{figure}
\end{document}